\documentclass[runningheads]{llncs}

\usepackage[mobile]{eccv}
\usepackage{eccvabbrv}
\usepackage{graphicx}
\usepackage{booktabs}
\usepackage{multirow}
\usepackage{makecell}
\usepackage{array}
\usepackage{adjustbox}
\usepackage{caption}
\usepackage{bm}
\usepackage[accsupp]{axessibility}
\usepackage{hyperref}
\usepackage{orcidlink}

\hypersetup{hidelinks}

\newcommand{\papertitle}{HSD: Training-Free Acceleration for \\
Document Parsing Vision-Language Models with \\
Hierarchical Speculative Decoding}

\makeatletter
\def\@fnsymbol#1{\ensuremath{
  \ifcase#1\or
    *\or
    \dagger\or
    \ddagger\or
    \mathsection\or
    \mathparagraph\or
    \|\or
    **\or
    \dagger\dagger\or
    \ddagger\ddagger
  \else
    \@ctrerr
  \fi}}
\makeatother

\makeatletter
\def\maketitlesupplementary{
    \clearpage
    \onecolumn
    \begin{center}
        {\Large\bfseries\boldmath \papertitle\par}
        \vspace{0.8em}
        {\large\bfseries Supplementary Material\par}
        \vspace{1.2em}
    \end{center}
}
\makeatother

\begin{document}
\title{\papertitle}

\author{
    Wenhui Liao\textsuperscript{\rm 1,2}$^{\dagger}$,
    Hongliang Li\textsuperscript{\rm 1,2}$^{\dagger}$,
    Pengyu Xie\textsuperscript{\rm 2,4},
    Xinyu Cai\textsuperscript{\rm 2},\\
    Yufan Shen\textsuperscript{\rm 2},
    Yi Xin\textsuperscript{\rm 2},
    Qi Qin\textsuperscript{\rm 2},
    Shenglong Ye\textsuperscript{\rm 2},
    Tianbin Li\textsuperscript{\rm 2},\\
    Ming Hu\textsuperscript{\rm 2},
    Junjun He\textsuperscript{\rm 2},
    Yihao Liu\textsuperscript{\rm 2},
    Wenhai Wang\textsuperscript{\rm 2},
    Min Dou\textsuperscript{\rm 2},\\
    Bin Fu\textsuperscript{\rm 2,3}$^{*}$,
    Botian Shi\textsuperscript{\rm 2}$^{*}$,
    Yu Qiao\textsuperscript{\rm 2}$^{*}$,
    Lianwen Jin\textsuperscript{\rm 1}$^{*}$
}
\authorrunning{W.~Liao and H.~Li et al.}
\institute{
South China University of Technology, Guangzhou, China \\
\and
Shanghai Artificial Intelligence Laboratory, Shanghai, China\\
\and
Shenzhen Institute of Advanced Technology, CAS, Shenzhen, China\\
\and
Nanjing University, Nanjing, China\\
\email{\{eelwh,eehongliangli\}@mail.scut.edu.cn}\\
\email{\{fubin,shibotian,qiaoyu\}@pjlab.org.cn}\\
\email{eelwjin@scut.edu.cn}
}

\titlerunning{HSD}

\maketitle

\begin{abstract}
    Document parsing is a fundamental task in multimodal understanding, supporting a wide range of downstream applications such as information extraction and intelligent document analysis. Benefiting from strong semantic modeling and robust generalization, VLM-based end-to-end approaches have emerged as the mainstream paradigm in recent years.
    However, these models often suffer from substantial inference latency, as they must autoregressively generate long, full-page sequences when processing long-form documents.
    While recent hybrid methods mitigate this issue via region-level parallel decoding with VLMs, independent region decoding loses full-page context and might weaken global coherence.
    To address this issue, we propose \textbf{Hierarchical Speculative Decoding (HSD)}, a two-stage local-to-global framework for document parsing.
    HSD first employs a lightweight pipeline drafter to predict region partitions and 
    generate coarse drafts for each region. 
    The first stage verifies the generated region-level drafts in parallel for efficiency, while the second stage further performs page-level verification on these refined outputs to preserve full-page coherence. 
    Experimental results show that our HSD achieves a near-lossless 2.78$\times$ speedup with HunyuanOCR on OmniDocBench v1.5 and up to 7.04$\times$ speedup on long-document parsing tasks, demonstrating the effectiveness of our proposed method. The code is available at https://github.com/whlscut/HSD.

\end{abstract}
\renewcommand{\thefootnote}{\fnsymbol{footnote}}
\footnotetext[2]{Equal contribution.}
\footnotetext[1]{Co-corresponding authors.}
\renewcommand{\thefootnote}{\arabic{footnote}}

\section{Introduction}
\label{sec:intro}

Document parsing~\cite{zhang2024document,nougat,mistral2025} converts document images into structured text by organizing text, formulas, tables, and figures in their natural reading order. 
As a foundational technology, it underpins a broad spectrum of downstream applications, including document indexing and retrieval, workflow automation, data governance, and large-scale corpus construction.
To enable these applications at scale, document parsing systems are expected to process massive volumes of documents with high accuracy and efficiency. However, achieving both objectives simultaneously remains challenging in practice.

Recent progress in document parsing can be divided into three categories: pipeline-based~\cite{marker,mineru,docling}, end-to-end~\cite{nougat, got,olmocr}, and hybrid~\cite{dolphin,monkeyocr,mineru2_5} approaches.
Pipeline-based methods decompose the workflow into sub-tasks (e.g., layout analysis~\cite{lggpt2025zhang}, reading-order prediction, text recognition, formula recognition, and table parsing), each handled by a lightweight specialist. This design enables region-level parallel recognition after layout segmentation, thereby improving efficiency. However, the overall performance is often limited by lightweight recognizers, and errors can propagate across stages.
End-to-end approaches are typically built on vision–language models (VLMs)~\cite{alayrac2022flamingo,li2022blip,chen2022pali,liu2023improved}, which directly generate the full parsing results for the input image. Benefiting from VLMs’ strong semantic modeling and the global coherence maintained during generation, these methods are usually more robust to complex layouts, noise, and cross-region dependencies.
However, autoregressive decoding makes long outputs inherently slow, causing latency to grow roughly linearly with sequence length.
Hybrid approaches aim to combine both paradigms: they first perform layout analysis to segment a page into semantic regions, and then decode these regions independently in parallel with the VLM. While region parallelism improves efficiency, independence across each individual region removes cross-region correlation (e.g., reading order, cross-column connections), which might damage page-level coherence.
Moreover, once the predicted layout or reading-order priors are wrong, the VLM is forced to decode under incorrect partitions/orders, resulting in error propagation.
In summary, existing approaches still struggle to simultaneously exploit region-level parallelism for efficiency while preserving page-level global coherence for robustness.

In this work, we show that the conflict between region-level parallelism and page-level global coherence is not inherent. To this end, we introduce Hierarchical Speculative Decoding (HSD), a new paradigm for end-to-end document parsing that performs two-stage speculative verification: coarse drafts are first verified in parallel at the region level, and then the refined outputs are verified globally at the page level to restore full-page coherence.
This design is motivated by the structured layout of documents, where content can be segmented into semantic regions such as paragraphs, tables, and figures.
Concretely, a lightweight pipeline model first produces a semantic region partition together with coarse predictions for each region,   which serve as speculative drafts.
In the region-level verification stage (Stage 1), these drafts are verified in parallel by the end-to-end parser.
However, since this stage lacks full-page context and may inherit layout segmentation errors from the pipeline parser, it will introduce structural inconsistencies such as incorrect layout hierarchy or reading order.
To address this problem, the page-level verification stage (Stage 2) performs full-page verification conditioned on the outputs of Stage 1.
Since these outputs have already been refined, page-level verification requires only a moderate number of steps to modify the remaining structural errors.

To further speed up the verification, we propose Decoupled Speculative Verification (DSV) for HSD. 
Unlike traditional speculative decoding, which repeatedly refreshes draft tokens to maintain prefix synchronization, we directly reuse the pipeline’s region predictions generated in a single forward pass as drafts.
While
it greatly reduces draft generation cost, 
such
decoupling introduces prefix–draft misalignment between the pre-generated drafts and the VLM’s current generation prefix. 
DSV utilizes a \textit{draft–target matching process} to address this alignment problem, and the \textit{prefix-tree batching mechanism} further enables efficient verification over multiple candidate matched segments.

To demonstrate the effectiveness of our method, we evaluate HSD on OmniDocBench v1.5~\cite{mineru2_5}, olmOCR-Bench~\cite{olmocr}, and Ocean-OCR-Bench~\cite{oceanocr} using multiple end-to-end parsers, including specialized document VLM parsers (dots. ocr~\cite{dots.ocr}, HunyuanOCR~\cite{hunyuanocr2025tencent}) and general-purpose VLMs (Qwen2.5-VL-3B/7B~\cite{qwen2.5-vl} and Qwen3-VL-2B/8B~\cite{qwen3-vl}). 
Experimental results demonstrate that our method delivers near-lossless acceleration across models, document types, and languages.
In particular, with HunyuanOCR, our method achieves end-to-end speedups of 2.78$\times$, 2.46$\times$, and 3.29$\times$ across OmniDocBench v1.5, olmOCR-Bench, and Ocean-OCR-Bench, respectively.

Our main contributions are threefold:
\begin{itemize}
\item We introduce \textbf{Hierarchical Speculative Decoding} for end-to-end document parsing, a paradigm that exploits {region-wise parallel verification}, and restores {global coherence} via {page-level verification}.
\item We propose \textbf{Decoupled Speculative Verification} for further acceleration, introducing a {draft–target matching process} to resolve misalignment and the {prefix-tree batching mechanism} to efficiently verify multiple draft segments.
\item Experiments across parsers and benchmarks show that our \textbf{training-free, plug-and-play} method achieves up to 7.04$\times$ speedup with near-lossless accuracy, highlighting a promising approach for efficient document parsing.
\end{itemize}

\section{Related Work}
\label{sec:related}

\subsection{Document Parsing}

\textbf{Traditional Pipelines.}
Document parsing~\cite{nanonets2025,marker,mineru} is inherently heterogeneous: pages mix text, mathematical expressions, tables, charts, and figures arranged in widely varying layouts and reading orders. To address these challenges, early systems decomposed the task into submodules—layout analysis, reading-order estimation, text recognition, formula recognition, table recognition, and final assembly. Representative toolkits such as Marker~\cite{marker}, MinerU~\cite{mineru}, PP-StructureV3~\cite{paddleocr3_0}, and Docling~\cite{docling} exemplify this design. For example, MinerU~\cite{mineru} begins with layout detection to partition a page into semantically labeled regions. It then routes each region to task-specific recognizers (text, equations, tables) and reconstructs the reading order to produce a consolidated Markdown result. These pipelines deliver practical throughput and engineering flexibility: lightweight specialists process layout regions in parallel, and modules can be swapped without retraining. However, cross-stage error propagation and limited submodule robustness can lead to failures on complex documents.

\textbf{End-to-End Approaches.}
Recent work~\cite{liu2025points,OCRFlux2025} has shifted toward end-to-end document parsing with a single autoregressive vision–language model (VLM). The model ingests a page image and directly emits structured markup, jointly modeling text, tables, equations, and reading order under a unified objective and long-context decoding. Early efforts such as Nougat~\cite{nougat} targeted scientific articles, converting pages into lightweight LaTeX/Markdown-style markup. GOT~\cite{got} generalizes this paradigm to broader document types and richer elements (e.g., molecular formulas, sheet music, geometric shapes, and charts) while emphasizing efficiency via a highly compressed vision encoder paired with a 0.5B decoder. Subsequent models such as Ocean-OCR~\cite{oceanocr}, olmOCR~\cite{olmocr}, dots.ocr~\cite{dots.ocr}, and HunyuanOCR~\cite{hunyuanocr2025tencent} push accuracy by adopting native-resolution encoders and training on larger, higher-quality, and more diverse corpora. Beyond supervised learning, researchers have also begun to explore reinforcement learning and verifiable training signals: Infinity-Parser~\cite{infinity_parser} optimizes layout fidelity with rewards on edit distance, paragraph counts, and structural consistency, whereas olmOCR 2~\cite{olmocr_2} uses binary unit tests as a programmatic reward signal. More recently, DeepSeek-OCR~\cite{wei2025deepseekocr} reduces token budgets by compressing visual context into fewer vision tokens before text decoding, shortening sequences and latency. Overall, end-to-end parsers now achieve competitive parsing accuracy, yet autoregressive decoding over long outputs remains a key inference bottleneck.

\textbf{Hybrid Approaches.}
Hybrid parsers~\cite{dolphin,monkeyocr,mineru2_5,paddleocr-vl} combine the efficiency of pipelines with the semantic capacity of end-to-end models. They first run a lightweight layout stage to segment and order semantic regions on the page, then parse each region with a vision–language model, and finally stitch the results into full-page markup. Representative systems include MonkeyOCR~\cite{monkeyocr}, Dolphin~\cite{dolphin}, MinerU2.5~\cite{mineru2_5}, and PaddleOCR-VL~\cite{paddleocr-vl}. 
Such designs reduce the effective sequence length per decoding call and exploit concurrency, but token generation within each region remains autoregressive, so long regions can still be latency-dominant.
Moreover, these methods can be sensitive to the quality of the layout analysis stage: boundary, ordering, or granularity mismatches between region decomposition and generation may propagate to the final parsing, potentially affecting cross-region coherence (e.g., multi-column flow and table continuity).
Finally, the multi-stage pipeline introduces additional interfaces and objectives, making holistic end-to-end optimization more involved than in single-model parsers.

\subsection{Speculative Decoding}
Speculative decoding~\cite{leviathan2023fast,chen2023acceleratinglargelanguagemodel,sun2021instantaneous,zhao2024ouroboros} accelerates autoregressive generation with a draft–verify scheme: a fast drafter proposes multiple next tokens, and the target model verifies them in a single forward pass using rejection sampling, preserving the target distribution while reducing sequential steps. Existing methods can be broadly classified along two axes. The first axis is who drafts: external drafters use a separate smaller model; internal drafters augment the target model with lightweight drafting modules, including Medusa~\cite{medusa}, EAGLE~\cite{eagle}, and EAGLE-2~\cite{eagle-2}; self-speculative methods~\cite{layerskip}, also called early-exit, allow the same network to draft with truncated layers and then verify with the full stack. The second axis concerns what is drafted: linear chunks or a branching tree~\cite{specinfer}. In linear drafting, the drafter predicts a single $k$-token continuation, and the verifier accepts the longest matching prefix before the first mismatch. In tree drafting~\cite{specinfer}, the drafter explores multiple alternatives at one or more steps, forming a token tree that is verified in parallel; this typically increases per-round acceptance, especially for long outputs, at the cost of additional draft compute and memory.
Extensions to VLMs~\cite{xie2026hivis,gagrani2024speculative,lin2025speculative,kang2025vispec,zhang2026sparrow} adapt these ideas to visual inputs and report around twofold throughput gains on systems such as LLaVA while maintaining quality.
Despite this progress, speculative decoding remains underexplored for document parsing, where outputs are typically long and highly structured.

\section{Methodology}
\label{sec:method}

\begin{figure*}[t]
  \centering
   \includegraphics[width=0.99\linewidth]{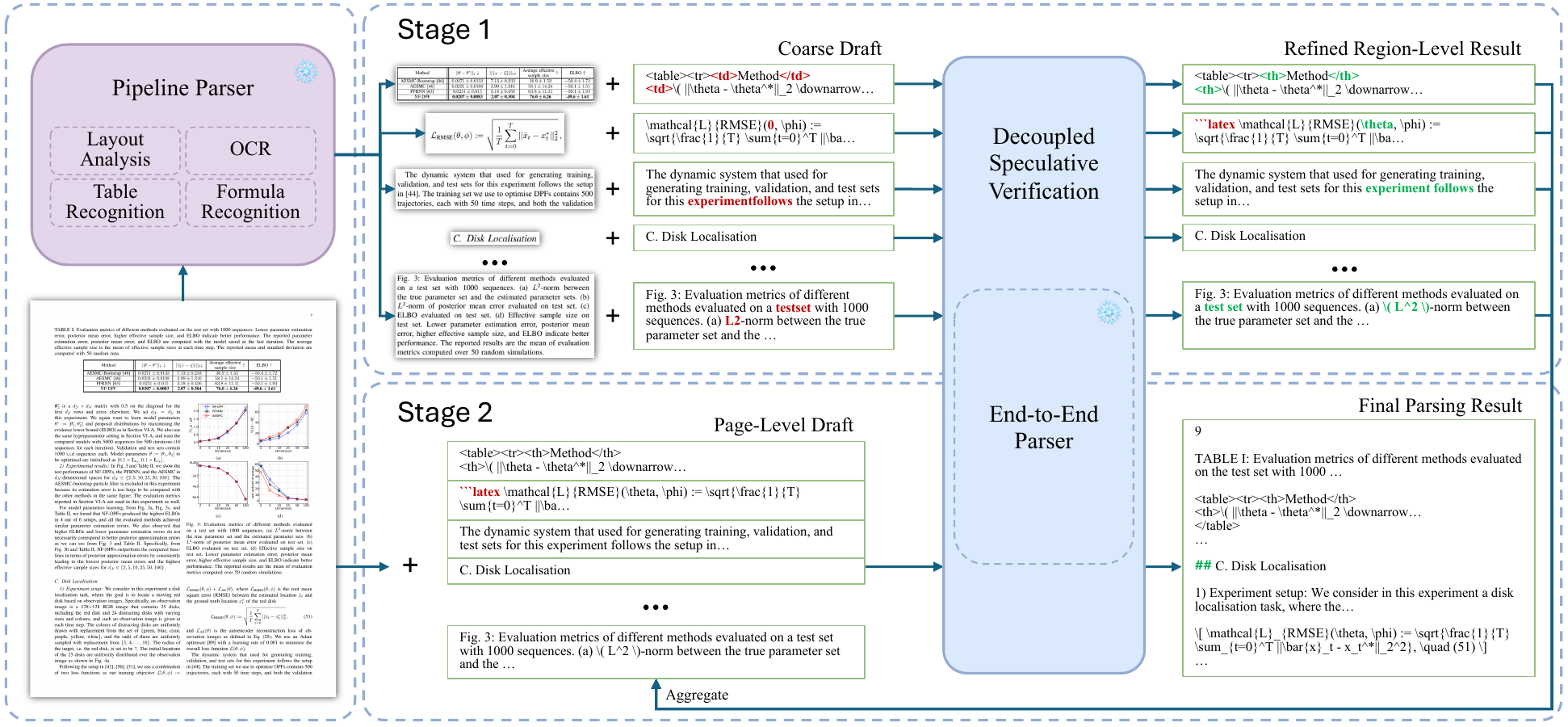}
   \caption{
Overview of the proposed hierarchical speculative decoding paradigm. A lightweight pipeline generates coarse region drafts via layout analysis and element recognition. The end-to-end parser verifies these drafts in two stages: Stage~1 performs region-level verification on cropped regions in parallel to produce refined drafts; Stage~2 aggregates them into a page-level draft and performs global, page-level verification to produce the final parsing result. \textcolor{BrickRed}{Red} and \textcolor{ForestGreen}{green} text denote draft errors and their corrections during verification, respectively.}
  \label{fig:main}
\end{figure*}

Hierarchical Speculative Decoding (HSD) is a training-free, inference-time acceleration method for end-to-end document parsers.
As shown in~\cref{fig:main},
a light-\allowbreak weight document parsing pipeline first performs layout analysis and element recognition (e.g., text/table/formula) to construct coarse drafts for document regions. The end-to-end parser then verifies these drafts hierarchically in two stages: (1) a region-level local verification stage that verifies drafts on cropped regions in parallel, producing refined region-level parses that are aggregated into a page draft; (2) a page-level global verification stage that verifies the page draft with modest multi-token decoding steps to obtain the final page-level parsing. In the following, we first introduce this two\mbox{-}stage hierarchical paradigm in Section~\ref{sec:hier}, and then provide details of our verification operator in Section~\ref{sec:specdecode}.

\subsection{Hierarchical Speculative Decoding Paradigm}
\label{sec:hier}

\paragraph{\textbf{Setup and Notation.}} Given the page image \(x\), the end-to-end parser \(p_\theta\) autoregressively produces tokens \(y_t\) forming a sequence \(\mathbf y = (y_{1:T})\) that spans text, formulas, tables, and figure markers, with conditional probabilities \(p_\theta(y_t | x, y_{<t})\). 
A lightweight pipeline \(q_\phi\) runs once per page and outputs a page layout \(\mathcal{R}=\{r_i\}_{i=1}^M\) with a set of fixed drafts \(\tilde{\mathcal Y}^{(i)}\)
for each region \(r_i\).
We denote the verification operator by \(\mathrm{SpecDecode}\):
\begin{equation}
\hat{\mathbf y}=\mathrm{SpecDecode}\big(p_\theta,\, z,\, \tilde{\mathcal Y}\big),
\end{equation}
which takes the current visual input \(z\) (either a region crop or the full page) along with a draft set \(\tilde{\mathcal Y}\), and returns a verified sequence \(\hat{\mathbf y}\).
\(\mathrm{SpecDecode}\) treats drafts as proposals and uses the document parser \(p_\theta\) to accept or correct them in several forward passes (details in Section~\ref{sec:specdecode}). 
Building on this operator, we instantiate a two-stage hierarchy in which the first stage verifies region drafts in parallel, and the second stage performs a single page-level pass to reconcile context.

\paragraph{\textbf{Stage~1 (Region-level Local Verification)}.}
For each \(r_i\in\mathcal R\), let \(z_i=\left.x\right|_{r_i}\) denote the cropped region. We verify the corresponding region drafts in parallel:
\begin{equation}
\hat{\mathbf y}^{(i)}=\mathrm{SpecDecode}\big(p_\theta,\, z_i,\, \tilde{\mathcal Y}^{(i)}\big).
\end{equation}
Parallel region-wise verification provides high throughput. 
However, since this stage lacks full-page context and may inherit layout segmentation errors from the pipeline parser, it can introduce structural inconsistencies, such as incorrect layout hierarchy or reading order.

\paragraph{\textbf{Stage~2 (Page-level Global Verification).}}
To address these residual errors, we aggregate Stage~1 outputs into an unordered collection as a page-level draft, 
\begin{equation}
\tilde{\mathcal{Y}}^{\mathrm{pg}}
= \left\{\, \hat{\mathbf{y}}^{(i)} \mid r_i \in \mathcal{R} \,\right\}, 
\end{equation}
which is also kept fixed during verification.

Then, we perform a full-page verification on the document:
\begin{equation}\label{eq:page}
\hat{\mathbf y}^{\text{pg}} = \mathrm{SpecDecode}\!\big(p_\theta, x, \tilde{\mathcal{Y}}^{\text{pg}}\big).
\end{equation}
The final reading order is resolved by \(p_\theta\) during verification. In other words, Stage~1 outputs serve as high-quality page-level drafts, allowing Stage~2 to complete verification in fewer forward steps.

\begin{figure}[t]
  \centering
   \includegraphics[width=0.95\linewidth]{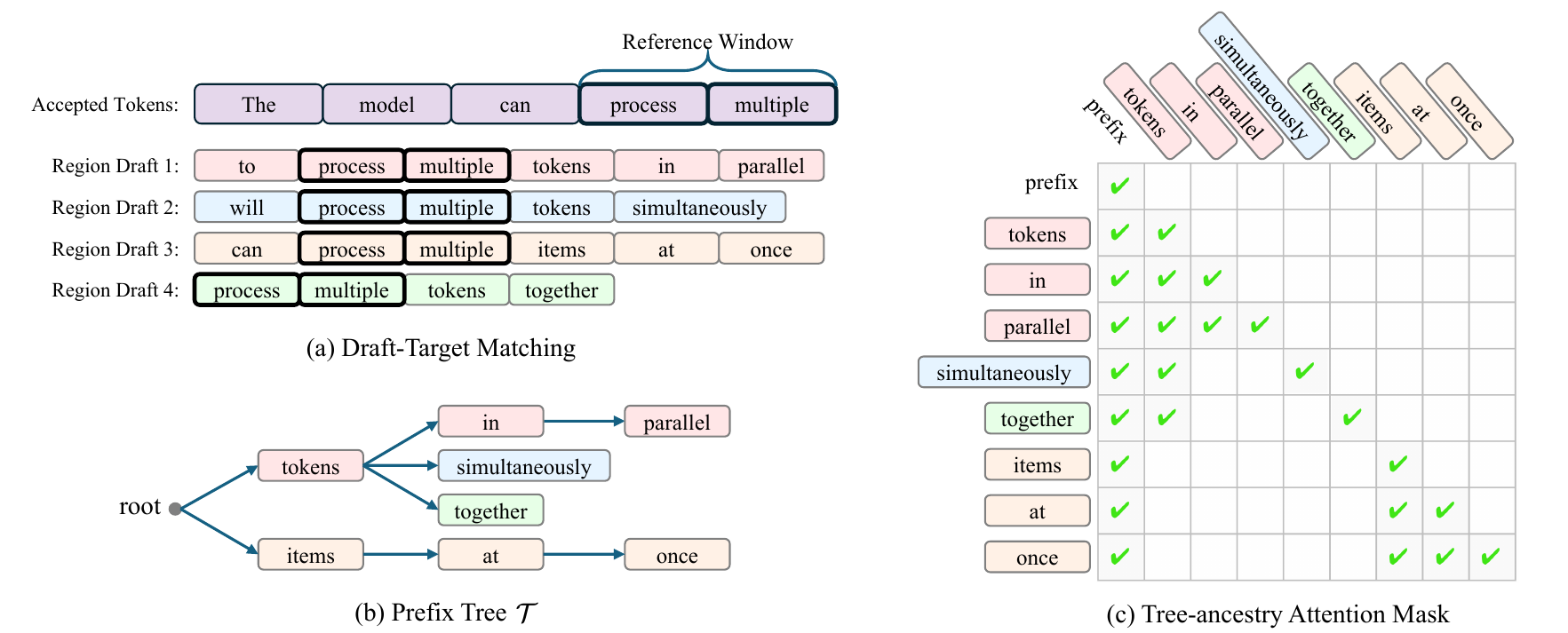}
   \caption{Visualization of decoupled speculative verification.
(a) Draft–target matching aligns a reference window from the accepted tokens with multiple drafts to extract candidate continuations.
(b) The prefix tree organizes candidates by merging common prefixes.
(c) The tree-ancestry attention mask enables parallel verification, where tokens attend only to the accepted tokens and their ancestors in the prefix tree (green checkmarks indicate allowed attention).
}
  \label{fig:dsv}
\end{figure}

\subsection{Decoupled Speculative Verification}
\label{sec:specdecode}

Traditional speculative decoding refreshes drafts synchronously at each decoding step, so the draft tokens are conditioned on the currently accepted tokens of the target model.
In contrast, our setting is \emph{decoupled}: a lightweight pipeline generates the coarse drafts \emph{once per page}, which remain unchanged during verification.
This decoupling introduces misalignments between the pre-generated drafts and the end-to-end parser’s current generation. 
To address this challenge, we design a two-part mechanism: (i) align drafts to the accepted tokens using a short \emph{reference window}, and (ii) verify \emph{multiple} candidates in one forward pass via a \emph{prefix-tree} formulation with a specialized attention mask.
\Cref{fig:dsv} provides an overall visualization
of the core mechanism, and we formalize it in the remainder of this section.

\paragraph{\textbf{Preliminaries.}} For any sequence \(a\), the slice notation \(a_{p:q}\) denotes the contiguous subsequence \((a_p,\ldots,a_q)\), and \(|a|\) denotes its length (in tokens).

\paragraph{\textbf{Draft–target matching process.}}
At decoding step \(t\), let \(\hat{\mathbf y}_{1:t}\) denote the accepted token sequence of the target model.
Let \(n\) denote the desired window length. As shown in \cref{fig:dsv}(a), we take a reference window to be the most recent \(n\) tokens of the accepted token sequence. Thus, the reference window can be expressed as \(\mathbf w=\hat{\mathbf y}_{t-n+1:t}\).

We slide the token sequence of the reference window \(\mathbf{w}\) across the draft \(\tilde{\mathbf{y}}\) and record every start position where it matches, forming \(\mathcal{J}(\tilde{\mathbf{y}})\):
{\small
\begin{equation}
\mathcal J(\tilde{\mathbf y})
=\bigl\{\, j\ \bigm|\ \tilde{\mathbf y}_{j:j+n-1}=\mathbf w\ \text{ and }\ 1\le j\le |\tilde{\mathbf y}|-n+1 \,\bigr\},
\end{equation}}
where \(j\) indexes the start position of each match.
We then extract the suffixes strictly following each matched window and collect them across all drafts in \(\tilde{\mathcal Y}\):
{\small
\begin{equation}
\mathcal C
=\Bigl\{\, \tilde{\mathbf y}_{\,j+n:|\tilde{\mathbf y}|}\ \Bigm|\ \tilde{\mathbf y}\in\tilde{\mathcal Y}\ \text{ and }\ j\in\mathcal J(\tilde{\mathbf y})\ \text{ and }\ j+n\le |\tilde{\mathbf y}| \Bigr\}.
\end{equation}
}
Through the above matching and extraction steps, we obtain a set of candidate suffixes $\mathcal{C}$ for verification in the subsequent stage.

\paragraph{\textbf{Prefix-tree batching mechanism.}}
When the number of candidates $|\mathcal{C}|>1$, verifying each candidate independently is redundant. 
To address this issue, we organize $\mathcal{C}$ into a prefix tree $\mathcal{T}$ that merges common prefixes, thereby enabling parallel verification. 
As shown in \cref{fig:dsv}(b), each node $v$ in $\mathcal{T}$ represents a unique prefix. 
Let $\pi(v)$ denote the token sequence along the path from the root to $v$. For example, if the node \(v\) is \texttt{parallel} (in \cref{fig:dsv}(b)), then \(\pi(v)=\texttt{tokens in parallel}\).

For any node $v$, we define the set of possible next tokens $\mathrm{Next}(v)$ as
{\small
\begin{equation}
\mathrm{Next}(v)=\Bigl\{\, \mathbf c_{|\pi(v)|+1}\; \Bigm|\;
\mathbf c\in\mathcal C\ \text{ and }\ \mathbf c_{1:|\pi(v)|}=\pi(v) \Bigr\}.
\end{equation}
}
Among all sequences $\mathbf c\in\mathcal C$ that share the prefix $\pi(v)$, 
$\mathrm{Next}(v)$ collects the distinct tokens that appear immediately after this prefix.
In \cref{fig:dsv}(b), for the node corresponding to the prefix \texttt{tokens}, we have \(\mathrm{Next}(v)=\{\texttt{in},\ \texttt{simultaneously},\ \\ 
\texttt{together}\}\).
For each token \(u\in\mathrm{Next}(v)\), there exists a unique node \(w\) with \(\pi(w)=\pi(v)\oplus u\), where $\oplus$ denotes concatenation.
We record \(w\) as \(\mathrm{child}(v,u)\), and then create a directed edge from \(v\) to \(\mathrm{child}(v,u)\).
In other words, each child node \(w\) encodes not only the token \(u\) itself, but also its prefix context \(\pi(v)\), thereby specifying the occurrence of \(u\) within a candidate sequence.

With the above definition, we construct the prefix tree recursively.
Starting from the root, which represents the empty prefix \((\pi(\mathrm{root})=\emptyset)\), we expand newly created children until every sequence in \(\mathcal C\) corresponds to a root-to-leaf path.
The resulting tree compactly encodes all candidate continuations by sharing common prefixes as shown in~\cref{fig:dsv}(b).

To enable parallel verification, we linearize \(\mathcal{T}\) into a packed sequence \(\mathcal{P}\) and apply a \emph{tree-ancestry attention mask}: a token at node \(v\) only attends to the accepted token sequence \(\hat{\mathbf{y}}_{1:t}\) and tokens along \(v\)’s ancestor path (see \cref{fig:dsv}(c)). This processes all candidate paths in one pass while preserving autoregressive conditioning.

\paragraph{\textbf{Verification and acceptance.}}
Let \(\tau\in(0,1)\) be the acceptance threshold. For any node \(v\), the model produces a next-token distribution
\begin{equation}
p_\theta(\cdot \mid z,\, \hat{\mathbf{y}}_{1:t} \oplus \pi(v)), 
\end{equation}
where \(z\) is the current visual input (a region crop or the full page).
We perform a greedy traversal of the prefix tree to obtain the final accepted token sequence.

At each step,
we select the most probable next token among the candidate set \(\mathrm{Next}(s)\):
\begin{equation}
u^\star=\arg\max_{u\in \mathrm{Next}(s)} p_\theta\!\left(u \mid z,\, \hat{\mathbf y}_{1:t}\oplus \pi(s)\right),
\end{equation}
where \(s\) is the current node.
Let \(\mathcal V\) denote the vocabulary. We define \(\hat u\in\mathcal V\) to be the token with the highest model probability under the current context:
\begin{equation}
\hat u=\arg\max_{u\in \mathcal V} p_\theta\!\left(u \mid z,\, \hat{\mathbf y}_{1:t}\oplus \pi(s)\right).
\end{equation}
We accept \(u^\star\) and move to its child node if
\begin{equation}
\log p_\theta\!\left(u^\star \mid z,\, \hat{\mathbf y}_{1:t}\oplus \pi(s)\right)
     - \log p_\theta\!\left(\hat u \mid z,\, \hat{\mathbf y}_{1:t}\oplus \pi(s)\right)
\;\ge\; \log \tau.
\end{equation}
If the condition fails or \(\mathrm{Next}(s) = \emptyset\), we stop at the current node \(s\).
Additionally, if \(s\) is a leaf node, there is no admissible next token and the traversal stops at \(s\).
Upon termination, we update the accepted token sequence:
\begin{equation}
\hat{\mathbf y}_{1:t_{\text{new}}} = \hat{\mathbf y}_{1:t} \oplus \pi(s) \oplus \hat u.
\end{equation}

By organizing candidates into a tree and verifying them in parallel, each step can accept multiple tokens at once, substantially reducing the number of decoding steps. Tree-structured batching increases per-step compute utilization without elongating the latency-critical path, yielding significant wall-clock speedup. The decoupled design leverages the high-throughput pipeline while keeping the target model as the arbiter that corrects draft errors.

\section{Experiments}
\label{sec:experiment}

\subsection{Datasets}
We evaluate on three public benchmarks: OmniDocBench v1.5~\cite{mineru2_5}, olmOCR-Bench~\cite{olmocr}, and Ocean-OCR-Bench~\cite{oceanocr}.
OmniDocBench v1.5~\cite{mineru2_5} contains 1,355 PDF pages spanning nine document types and provides rich evaluation for document parsing, including 15 block-level categories and 4 span-level elements with text, LaTeX formulas, tables, reading-order annotations, and page/block attributes. We follow the official end-to-end evaluation setup.
olmOCR-Bench~\cite{olmocr} includes 1,403 PDFs paired with 7,010 unit tests that check properties of PDF-to-Markdown conversion, such as content presence, natural reading order, table fidelity, and mathematical expressions. We report the official aggregate score under its evaluation protocol. 
Ocean-OCR-Bench~\cite{oceanocr} is a bilingual page-level evaluation built from 200 document images (100 English and 100 Chinese). The official metrics include normalized edit distance, F1, precision, recall, BLEU, and METEOR; we adopt the same protocol across different settings.

\subsection{Metrics}
For a comprehensive evaluation, we report three efficiency-related metrics: Decoding Speedup, End-to-End Speedup, and Average Acceptance Length.

\textbf{Decoding Speedup~\cite{zhang2026sparrow}.} Decoding Speedup measures the acceleration of the latency-critical generation loop:
\begin{equation}
\mathrm{SR}_{\text{decode}}=\frac{T^{\mathrm{AR}}_{\mathrm{decode}}}{T^{\mathrm{Spec}}_{\mathrm{decode}}}.
\end{equation}
Here, \(T^{\mathrm{AR}}_{\mathrm{decode}}\) and \(T^{\mathrm{Spec}}_{\mathrm{decode}}\) are wall-clock times for standard autoregressive decoding and our speculative decoder, respectively. Unless otherwise noted, \(T_{\mathrm{decode}}\) \emph{includes} the draft model’s forward passes used for speculation, target-model parallel verification, accept/reject control flow (including rollbacks), KV-cache maintenance, and communication overheads; it \emph{excludes} disk I/O, non-generative preprocessing, and prefill not executed inside the decoding loop.

\textbf{End-to-End Speedup~\cite{zhang2026sparrow}.} To reflect user-perceived latency, we report end-to-end Speedup:
\begin{equation}
\mathrm{SR}_{\text{e2e}}=\frac{T^{\mathrm{AR}}_{\mathrm{full}}}{T^{\mathrm{Spec}}_{\mathrm{full}}},
\end{equation}
where \(T_{\mathrm{full}}\) measures the time from page image input to final parsing result,
\emph{including} vision/prefill computation (e.g., visual encoder and text prefill), while excluding disk I/O. Draft generation triggered specifically for speculation (e.g., per-region drafts) is counted in both \(\mathrm{SR}_{\text{decode}}\) and \(\mathrm{SR}_{\text{e2e}}\).

\textbf{Average Acceptance Length~\cite{leviathan2023fast}.} To quantify how many decoding steps are saved by speculation, we report Average Acceptance Length (AAL). For verification step \(k\), let \(\alpha_k\) denote the number of consecutive draft tokens accepted by the target model before the first mismatch (full rejection gives \(\alpha_k=0\)). Then,

\begin{equation}
\mathrm{AAL}=\frac{1}{N}\sum_{k=1}^{N} \alpha_k,
\end{equation}
with \(N\) being the number of verification steps. Larger AAL indicates more tokens skipped per step and thus higher potential speedup, though the realized \(\mathrm{SR}\) also depends on per-step verification overhead and parallel efficiency.

\subsection{Implementation Details}
We validate the proposed acceleration method on several mainstream end-to-end parsers, including dots.ocr~\cite{dots.ocr}, HunyuanOCR~\cite{hunyuanocr2025tencent}, Qwen2.5-VL-3B~\cite{qwen2.5-vl}, Qwen2.5-VL-7B~\cite{qwen2.5-vl}, Qwen3-VL-2B~\cite{qwen3-vl}, and Qwen3-VL-8B~\cite{qwen3-vl}. All experiments are conducted on NVIDIA A100 GPUs. 
For fair comparison, all methods are evaluated using the same {Hugging Face} Transformers stack~\cite{Wolf2020transformers}, with FlexAttention~\cite{FlexAttention} enabled for attention computation.
Our method uses PP-StructureV3 \cite{paddleocr3_0} by default for layout analysis and region draft generation. 
In Decoupled Speculative Verification, we set the reference window length to $n=3$, 
the acceptance threshold to $\tau=0.75$, and the maximum length of the packed sequence \(\mathcal{P}\) to 128 tokens.

\begin{table*}[!t]
  \caption{Acceleration across different models on OmniDocBench v1.5. Results are reported for our proposed Hierarchical Speculative Decoding. AAL denotes the average number of accepted draft tokens per verification step; $SR_{\mathrm{decode}}$ is the decode-only speedup (draft generation + verification); $SR_{\mathrm{e2e}}$ is the end-to-end page-level latency speedup (including vision/prefill stages).}
  \label{tab:omni}
  \centering
  \begin{adjustbox}{width=0.95\textwidth}
  \begin{tabular}{lccccccccccccc}
    \toprule
    \multirow{2}*{Model}&\multirow{2}*{Parameters}&\multicolumn{3}{c}{Overall}&Slides&\makecell{Academic\\Papers}&Book&Textbook&\makecell{Exam\\Papers}&Magazine&Newspaper&Notes&\makecell{Financial\\Report}\\

    \cmidrule(lr){3-5} \cmidrule(lr){6-6} \cmidrule(lr){7-7} \cmidrule(lr){8-8} \cmidrule(lr){9-9}\cmidrule(lr){10-10}\cmidrule(lr){11-11}\cmidrule(lr){12-12}\cmidrule(lr){13-13}\cmidrule(lr){14-14}

   \multicolumn{2}{c}{}&\multicolumn{1}{c}{AAL}&\multicolumn{1}{c}{$SR_{decode}$}&\multicolumn{1}{c}{$SR_{e2e}$}&\multicolumn{1}{c}{$SR_{e2e}$}&\multicolumn{1}{c}{$SR_{e2e}$}&\multicolumn{1}{c}{$SR_{e2e}$}&\multicolumn{1}{c}{$SR_{e2e}$}&\multicolumn{1}{c}{$SR_{e2e}$}&\multicolumn{1}{c}{$SR_{e2e}$}&\multicolumn{1}{c}{$SR_{e2e}$}&\multicolumn{1}{c}{$SR_{e2e}$}&\multicolumn{1}{c}{$SR_{e2e}$} \\

    \midrule
    Qwen2.5-VL-7B~\cite{qwen2.5-vl} & 8B & 3.56 & 2.13$\times$ & 2.10$\times$ & 1.43$\times$ & 2.35$\times$ & 1.99$\times$ & 1.73$\times$ & 1.77$\times$ & 2.17$\times$ & 2.95$\times$ & 1.78$\times$ & 1.94$\times$ \\
    Qwen2.5-VL-3B~\cite{qwen2.5-vl} & 4B & 2.52 & 2.14$\times$ & 2.12$\times$ & 1.93$\times$ & 2.37$\times$ & 2.02$\times$ & 1.80$\times$ & 1.77$\times$ & 2.39$\times$ & 2.80$\times$ & 1.99$\times$ & 2.19$\times$  \\
    Qwen3-VL-8B~\cite{qwen3-vl} &9B& 3.98 & 2.62$\times$ & 2.61$\times$ & 1.63$\times$ & 2.56$\times$ & 2.29$\times$ & 2.02$\times$ & 2.18$\times$ & 2.59$\times$ & 4.62$\times$ & 1.86$\times$ & 1.91$\times$ \\
    Qwen3-VL-2B~\cite{qwen3-vl} &2B& 3.33 & 2.20$\times$ & 2.18$\times$ & 1.35$\times$ & 2.13$\times$ & 1.82$\times$ & 1.87$\times$ & 2.04$\times$ &  2.61$\times$ & 4.03$\times$ & 1.69$\times$ & 1.75$\times$ \\
    dots.ocr~\cite{dots.ocr} &3B & 3.98 & 2.44$\times$ & 2.42$\times$ & 1.52$\times$ & 3.47$\times$ & 2.28$\times$ & 2.29$\times$  & 2.04$\times$ & 2.34$\times$ & 2.98$\times$ & 1.39$\times$ & 4.89$\times$\\
    HunyuanOCR~\cite{hunyuanocr2025tencent} & 0.9B & 4.55 & 2.82$\times$ & 2.78$\times$ & 1.58$\times$ & 3.41$\times$ & 4.00$\times$ & 1.92$\times$ & 1.72$\times$ & 2.60$\times$ & 4.30$\times$ & 1.98$\times$ & 7.04$\times$ \\
    
  \bottomrule
  \end{tabular}
  \end{adjustbox}
\end{table*}

\begin{table*}[!t]
  \caption{Acceleration across different models on olmOCR-Bench. Results are reported for our proposed Hierarchical Speculative Decoding.}
  \label{tab:olm}
  \centering
  \begin{adjustbox}{width=0.85\textwidth}
  \begin{tabular}{lccccccccccc}
    \toprule
    \multirow{2}*{Model}&\multirow{2}*{Parameters}&\multicolumn{3}{c}{Overall}&\makecell{arXiv\\Math}&\makecell{Old Scans\\Math}&Tables&Old Scans&\makecell{Headers\\Footers}&\makecell{Multi\\Column}&\makecell{Long Tiny\\Text}\\

    \cmidrule(lr){3-5} \cmidrule(lr){6-6} \cmidrule(lr){7-7} \cmidrule(lr){8-8} \cmidrule(lr){9-9}\cmidrule(lr){10-10}\cmidrule(lr){11-11}\cmidrule(lr){12-12}

   \multicolumn{2}{c}{}&\multicolumn{1}{c}{AAL}&\multicolumn{1}{c}{$SR_{decode}$}&\multicolumn{1}{c}{$SR_{e2e}$}&\multicolumn{1}{c}{$SR_{e2e}$}&\multicolumn{1}{c}{$SR_{e2e}$}&\multicolumn{1}{c}{$SR_{e2e}$}&\multicolumn{1}{c}{$SR_{e2e}$}&\multicolumn{1}{c}{$SR_{e2e}$}&\multicolumn{1}{c}{$SR_{e2e}$}&\multicolumn{1}{c}{$SR_{e2e}$} \\

    \midrule
    Qwen2.5-VL-7B~\cite{qwen2.5-vl} & 8B & 2.47 & 2.05$\times$ & 2.01$\times$ & 2.04$\times$ & 1.57$\times$ & 1.49$\times$ & 1.01$\times$ & 2.06$\times$ & 2.53$\times$ & 1.59$\times$ \\
    Qwen2.5-VL-3B~\cite{qwen2.5-vl} & 4B & 2.66 & 2.67$\times$ & 2.64$\times$ & 2.37$\times$ & 1.78$\times$ & 2.41$\times$ & 1.05$\times$ & 3.18$\times$ & 4.07$\times$ & 3.08$\times$ \\
    Qwen3-VL-8B~\cite{qwen3-vl} &9B& 3.08 & 2.42$\times$ & 2.40$\times$ & 2.51$\times$ & 1.64$\times$ & 1.60$\times$ & 1.11$\times$ & 2.65$\times$ & 3.75$\times$ & 2.19$\times$ \\
    Qwen3-VL-2B~\cite{qwen3-vl} &2B& 3.91 & 2.98$\times$ & 2.97$\times$ & 3.00$\times$ & 1.65$\times$ & 2.20$\times$ & 1.46$\times$ & 3.20$\times$  & 4.48$\times$ & 2.54$\times$ \\
    dots.ocr~\cite{dots.ocr} &3B& 3.03 & 2.30$\times$ & 2.27$\times$ & 2.11$\times$ & 2.10$\times$ & 2.45$\times$ & 1.58$\times$ & 2.70$\times$ & 2.11$\times$ & 2.61$\times$  \\
    HunyuanOCR~\cite{hunyuanocr2025tencent} & 0.9B & 3.54 & 2.50$\times$ & 2.46$\times$ & 2.03$\times$ & 1.40$\times$ & 3.77$\times$ & 1.28$\times$ & 3.06$\times$ & 3.72$\times$ & 1.93$\times$ \\

      \bottomrule
  \end{tabular}
  \end{adjustbox}
\end{table*}

\subsection{Comprehensive Evaluations and Comparisons}

\textbf{Acceleration Performance Across Diverse Document Types and Models.}
To comprehensively evaluate the effectiveness of our hierarchical speculative decoding (HSD) approach, we conduct experiments on three benchmarks (OmniDocBench v1.5, olmOCR-Bench, and Ocean-OCR-Bench) across multiple end-to-end parsers, including specialized document parsers (dots.ocr, HunyuanOCR) and general-purpose VLMs (Qwen2.5-VL-7B/3B, Qwen3-VL-8B/2B). As shown in \cref{tab:omni,tab:olm,tab:ocean}, HSD consistently delivers positive and substantial speedups across diverse models and benchmarks with no obvious slowdown. In particular, on the state-of-the-art end-to-end parser HunyuanOCR, we obtain end-to-end speedups of 2.78$\times$ (OmniDocBench v1.5), 2.46$\times$ (olmOCR-Bench), and 3.29$\times$ (Ocean-OCR-Bench), and observe similar trends on other parsers. The speedup magnitude varies by document type, mainly due to differences in output length, layout structure, and draft quality. In general, long documents with multiple semantic blocks (e.g., \emph{Newspaper} and \emph{Academic Papers} in OmniDocBench v1.5) offer higher region-level parallelism and thus larger gains, whereas challenging handwritten or degraded scans (e.g., \emph{Old Scans} in olmOCR-Bench) often yield lower-quality drafts, reducing acceptance and limiting acceleration. We provide qualitative visualizations in~\cref{sec:qualitative_analysis_speedup} to substantiate these observations.
Overall, HSD remains broadly effective across end-to-end parsers and document domains, demonstrating strong generality.

\begin{table}[!t]
  \caption{Acceleration performance across different models on Ocean-OCR-Bench. Results are reported for our proposed Hierarchical Speculative Decoding.}
  \label{tab:ocean}
  \centering
  \begin{adjustbox}{width=0.59\textwidth}
  \begin{tabular}{@{}lcccccc@{}}
    \toprule
    \multirow{2}*{Model}&\multirow{2}*{Parameters}&\multicolumn{3}{c}{Overall}&English&Chinese\\

    \cmidrule(lr){3-5} \cmidrule(lr){6-6} \cmidrule(lr){7-7}

   \multicolumn{2}{c}{}&\multicolumn{1}{c}{AAL}&\multicolumn{1}{c}{$SR_{decode}$}&\multicolumn{1}{c}{$SR_{e2e}$}&\multicolumn{1}{c}{$SR_{e2e}$}&\multicolumn{1}{c}{$SR_{e2e}$} \\
    \midrule
    Qwen2.5-VL-7B~\cite{qwen2.5-vl} & 8B & 4.91 & 3.03$\times$ & 3.00$\times$ & 3.62$\times$ & 2.62$\times$ \\
    Qwen2.5-VL-3B~\cite{qwen2.5-vl} & 4B & 2.63 & 2.78$\times$ & 2.72$\times$ & 4.51$\times$ & 2.04$\times$  \\
    Qwen3-VL-8B~\cite{qwen3-vl} &9B& 6.76 & 3.74$\times$ & 3.70$\times$ & 3.54$\times$ & 3.57$\times$ \\
    Qwen3-VL-2B~\cite{qwen3-vl} &2B& 3.11 & 3.02$\times$ & 2.99$\times$ & 4.81$\times$ & 2.07$\times$   \\
    dots.ocr~\cite{dots.ocr} &3B& 5.79 & 3.79$\times$ & 3.68$\times$ & 3.61$\times$ & 3.75$\times$ \\
    HunyuanOCR~\cite{hunyuanocr2025tencent} & 0.9B & 4.73 & 3.37$\times$ & 3.29$\times$ & 3.96$\times$ & 2.86$\times$ \\

  \bottomrule
  \end{tabular}
  \end{adjustbox}
\end{table}

\begin{table}[!t]
  \caption{Comparisons with speculative decoding baselines under the same target model.}
  \label{tab:compare_with_spec}
  \centering
  \begin{adjustbox}{width=0.85\linewidth}
  \begin{tabular}{@{}lllcccccc@{}}
    \toprule
    \multirow{2}{*}{Target Model}
    & \multirow{2}{*}{Method}
    & \multirow{2}{*}{Venue}
    & \multicolumn{2}{c}{OmniDocBench v1.5}
    & \multicolumn{2}{c}{olmOCR-Bench}
    & \multicolumn{2}{c}{Ocean-OCR-Bench} \\
    \cmidrule(lr){4-5} \cmidrule(lr){6-7} \cmidrule(lr){8-9}
    & & & AAL & $SR_{e2e}$ & AAL & $SR_{e2e}$ & AAL & $SR_{e2e}$ \\
    \midrule
    \multirow{12}{*}{Qwen2.5-VL-3B}
    & VSD~\cite{leviathan2023fast}
      & ICML 2023
      & 5.51 & 1.01$\times$
      & 5.62 & 1.02$\times$
      & 5.71 & 1.12$\times$ \\
    & Spec-LLaVA~\cite{huo2025specllava}
      & ICML 2025
      & 1.19 & 1.01$\times$
      & 1.23 & 1.02$\times$
      & 1.40 & 1.08$\times$ \\
    & SpecVLM~\cite{huang2025specvlm}
      & arXiv 2025
      & 1.89 & 1.07$\times$
      & 1.79 & 1.23$\times$
      & 1.87 & 1.18$\times$ \\
    & Medusa~\cite{medusa}
      & ICML 2024
      & 0.33 & 1.26$\times$
      & 0.31 & 1.28$\times$
      & 0.35 & 1.32$\times$ \\
    & MSD~\cite{lin2025speculative}
      & arXiv 2025
      & 0.81 & 1.30$\times$
      & 1.15 & 1.67$\times$
      & 1.10 & 1.65$\times$ \\
    & Dream~\cite{hu2026dream}
      & NeurIPS 2025
      & 2.86 & 1.46$\times$
      & 2.87 & 1.62$\times$
      & 3.10 & 1.53$\times$ \\
    & HiViS~\cite{xie2026hivis}
      & CVPR 2025
      & 1.04 & 1.56$\times$
      & 1.22 & 1.65$\times$
      & 1.44 & 1.72$\times$ \\
    & Sparrow~\cite{zhang2026sparrow}
      & arXiv 2026
      & 0.96 & 1.57$\times$
      & 0.99 & 1.53$\times$
      & 1.02 & 1.59$\times$ \\
    & EAGLE-2~\cite{eagle-2}
      & EMNLP 2024
      & 1.50 & 1.69$\times$
      & 1.56 & 1.86$\times$
      & 1.19 & 1.69$\times$ \\
    & ViSpec~\cite{kang2025vispec}
      & NeurIPS 2025
      & 1.64 & 1.75$\times$
      & 1.62 & 1.90$\times$
      & 1.22 & 1.72$\times$ \\
    & \textbf{HSD}
      & Ours
      & 2.52 & \textbf{2.12$\times$}
      & 2.66 & \textbf{2.64$\times$}
      & 2.63 & \textbf{2.72$\times$} \\
    \bottomrule
  \end{tabular}
  \end{adjustbox}
\end{table}

\textbf{Comparisons with Existing Speculative Decoding Baselines.}
We compare HSD with representative speculative decoding methods on document parsing benchmarks under the same target model.
Specifically, for vanilla speculative decoding (VSD)~\cite{leviathan2023fast}, we use a lightweight InternVL3.5-1B~\cite{wang2025internvl35} model as the drafter for the Qwen2.5-VL-3B target model, since no smaller model from the same family is available.
We further adapt Medusa~\cite{medusa} and EAGLE-2~\cite{eagle-2} to VLM-based document parsing by integrating their drafting modules into the target VLM.
For fairness, we also task-adapt the drafters of VSD, Medusa, EAGLE-2, and recent VLM-oriented speculative decoding methods, including Spec-LLaVA~\cite{huo2025specllava}, SpecVLM~\cite{huang2025specvlm}, MSD~\cite{lin2025speculative}, Dream~\cite{hu2026dream}, HiVis~\cite{xie2026hivis}, Sparrow~\cite{zhang2026sparrow}, and ViSpec~\cite{kang2025vispec}, using the same document parsing data.

As shown in \cref{tab:compare_with_spec}, existing speculative decoding methods provide only limited acceleration on document parsing, with the best baseline, ViSpec, achieving a speedup of 1.72--1.90$\times$.
This suggests that speculative decoding for document parsing poses a unique challenge.
In contrast, HSD achieves substantially higher speedups without additional training, reaching 2.12--2.72$\times$ under the same Qwen2.5-VL-3B target model.
This advantage comes from two designs tailored to document parsing: \textit{a pipeline parser that efficiently produces high-acceptance drafts, and a hierarchical framework that enables region-level parallel verification while preserving page-level parsing accuracy}.

\textbf{Comparisons with Existing Document Parsing Methods.}
We further compare end-to-end parsers~\cite{dots.ocr, hunyuanocr2025tencent} with and without HSD against representative pipeline-based~\cite{MinerUgithub, paddleocr3_0} and hybrid~\cite{monkeyocr, mineru2_5} document parsing approaches.
The comparison is intended to characterize the accuracy--latency trade-off of different parsing paradigms under their standard inference settings, rather than to perform a latency-matched evaluation.
As shown in~\cref{tab:pipeline/hybrid}, pure pipelines are fast but often result in lower output quality, while hybrid systems aim to strike a balance between quality and latency.
End-to-end parsers can achieve stronger quality on complex pages but are markedly slower.
By integrating HSD, we reduce the latency of end-to-end parsers by 2.42--2.78$\times$ with essentially unchanged quality, bringing them closer to the efficiency level of hybrid systems.
Overall, HSD establishes a new paradigm for document parsing, enabling end-to-end parsers to attain hybrid-level efficiency without altering the underlying models or introducing extra training.

\begin{table}[t]
  \caption{Comparison with pipeline, hybrid, and end-to-end document parsers on OmniDocBench v1.5, including end-to-end parsers with and without our HSD.}
  \label{tab:pipeline/hybrid}
  \centering
  \begin{adjustbox}{width=0.85\linewidth}
  \begin{tabular}{@{}lcccccccc@{}}
    \toprule
    \multirow{3}*{Model}&\multicolumn{2}{c}{Pipeline}&\multicolumn{2}{c}{Hybrid}&\multicolumn{2}{c}{End-to-End}&\multicolumn{2}{c}{\textbf{HSD}}\\

    \cmidrule(lr){2-3} \cmidrule(lr){4-5} \cmidrule(lr){6-7} \cmidrule(lr){8-9}

   \multicolumn{1}{c}{}&\multicolumn{1}{c}{\shortstack{MinerU2-\\pipeline~\cite{MinerUgithub}}}&\multicolumn{1}{c}{\shortstack{PP-Structure\\V3~\cite{paddleocr3_0}}}&\multicolumn{1}{c}{\shortstack{MonkeyOCR-\\pro-3B~\cite{monkeyocr}}}&\multicolumn{1}{c}{\shortstack{MinerU\\2.5~\cite{mineru2_5}}}&\multicolumn{1}{c}{\shortstack{dots.\\ocr~\cite{dots.ocr}}}&\multicolumn{1}{c}{\shortstack{Hunyuan\\OCR~\cite{hunyuanocr2025tencent}}}&\multicolumn{1}{c}{\shortstack{\textbf{dots.}\\\textbf{ocr}~\cite{dots.ocr}}}&\multicolumn{1}{c}{\shortstack{\textbf{Hunyuan}\\\textbf{OCR}~\cite{hunyuanocr2025tencent}}} \\

    \midrule
    Accuracy$\uparrow$ & 75.51 & 86.73 & 88.85 & 90.67 & 86.73 & 94.10 & 88.81 & 94.02  \\
    Mean latency (s/sample)$\downarrow$ & 1.79 & 1.99 & 26.69 & 46.74 & 60.06 & 30.47 & 24.82 & 10.96  \\
    
  \bottomrule
  \end{tabular}
  \end{adjustbox}
\end{table}

\begin{table}[!t]
  \caption{Acceleration when combining our hierarchical speculative decoding (HSD) with the visual token compression (VTC) method.}
  \label{tab:deepseek-ocr}
  \centering
  \begin{adjustbox}{width=0.75\textwidth}
  \begin{tabular}{lccccccccc}
    \toprule
    \multirow{2}*{Method}&\multicolumn{3}{c}{OmniDocBench v1.5}&\multicolumn{3}{c}{olmOCR-Bench}&\multicolumn{3}{c}{Ocean-OCR-Bench}\\

    \cmidrule(lr){2-4} \cmidrule(lr){5-7} \cmidrule(lr){8-10}

   \multicolumn{1}{c}{}&\multicolumn{1}{c}{AAL}&\multicolumn{1}{c}{$SR_{decode}$}&\multicolumn{1}{c}{$SR_{e2e}$}&\multicolumn{1}{c}{AAL}&\multicolumn{1}{c}{$SR_{decode}$}&\multicolumn{1}{c}{$SR_{e2e}$}&\multicolumn{1}{c}{AAL}&\multicolumn{1}{c}{$SR_{decode}$}&\multicolumn{1}{c}{$SR_{e2e}$} \\

    \midrule

    VTC (DeepSeek-OCR~\cite{wei2025deepseekocr}) & - & 1.00$\times$ & 1.00$\times$ & - & 1.00$\times$ & 1.00$\times$ & - & 1.00$\times$ & 1.00$\times$ \\
    + HSD & 3.72 & 1.51$\times$ & 1.56$\times$ & 3.65 & 1.39$\times$ & 1.41$\times$ & 5.36 & 1.87$\times$ & 1.91$\times$ \\
    
  \bottomrule
  \end{tabular}
  \end{adjustbox}
\end{table}

\textbf{Combining with the Visual Token Compression Method.}
Complementary to HSD, visual token compression (VTC) accelerates document parsing by reducing the number of visual tokens and thus the attention cost (e.g., DeepSeek-OCR~\cite{wei2025deepseekocr}). 
To examine whether the two directions are compatible, we integrate HSD with DeepSeek-OCR. 
As shown in~\cref{tab:deepseek-ocr}, HSD brings additional speedups on top of VTC, suggesting that our method is plug-and-play and stackable with other acceleration techniques.

\subsection{Ablation Study}

\begin{figure}[t]
  \centering

  \adjustbox{valign=t}{%
  \begin{minipage}[t]{0.49\linewidth}
    \centering

    \begingroup
      \setlength{\abovecaptionskip}{0pt}%
      \captionof{table}{Accuracy comparison of the draft model, baseline parser, and our hierarchically accelerated variants.}
      \label{tab:acc}
    \endgroup
    \begin{adjustbox}{width=0.92\linewidth}
      \begin{tabular}{@{}lcccc@{}}
        \toprule
        \multicolumn{2}{c}{Method} & \makecell{OmniDocBench\\v1.5} & \makecell{olmOCR-\\Bench} & \makecell{Ocean-OCR-\\Bench}  \\
        \midrule
        \multicolumn{2}{c}{Pipeline (Draft)} & 86.73 & 65.80 & 85.20 \\
        \midrule
        \multirow{3}*{\shortstack{dots.ocr\\~\cite{dots.ocr}}} & Baseline & 88.41 & 79.90 & 91.45 \\
        \multicolumn{1}{c}{} & Stage 1 only & 70.47 & 67.30 & 86.92 \\
        \multicolumn{1}{c}{} & Stage 1+2 (Ours) & 88.81 & 79.40 & 92.56 \\
        \bottomrule
      \end{tabular}
    \end{adjustbox}

    \captionof{table}{Ablation of our design on OmniDocBench v1.5 with dots.ocr.}
    \label{tab:ablation}
    \begin{adjustbox}{width=0.92\linewidth}
      \begin{tabular}{@{}lccc@{}}
        \toprule
        Method & AAL & $SR_{decode}$ & $SR_{e2e}$ \\
        \midrule
        Baseline & - & 1.00$\times$ & 1.00$\times$ \\
        + Page-level Spec. Decoding only & 2.49 & 2.11$\times$ & 2.09$\times$ \\
        + Hierarchical Spec. Decoding & 3.98 & 2.44$\times$ & 2.42$\times$ \\
        \bottomrule
      \end{tabular}
    \end{adjustbox}

  \end{minipage}%
  }\hfill
  \adjustbox{valign=t}{%
  \begin{minipage}[t]{0.485\linewidth}
    \centering

    \includegraphics[width=0.96\linewidth]{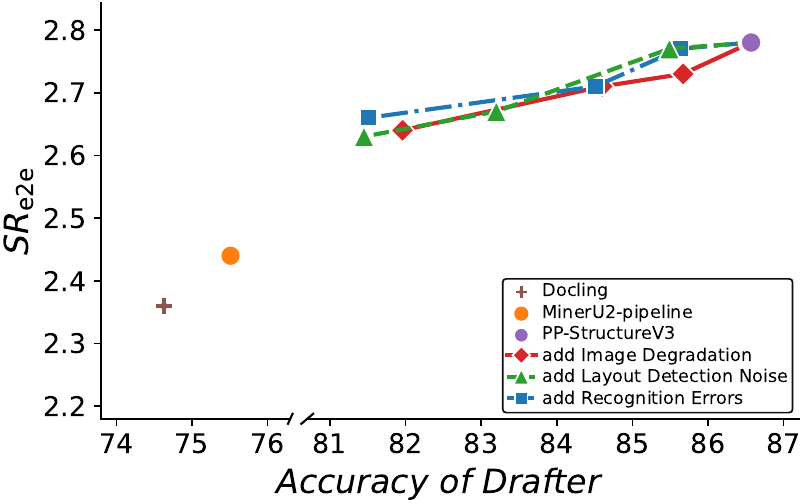}

    \caption{Impacts of using different pipelines as drafters. \textit{Accuracy of Drafter} is the OmniDocBench v1.5 score; ``add ...'' denotes noise injected into PP-StructureV3 drafts.}
    \label{fig:drafter_impact}
  \end{minipage}%
  }

\end{figure}

\textbf{Accuracy Analysis of Hierarchical Design.}
\cref{tab:acc} validates the effectiveness of our hierarchical design and demonstrates near-lossless acceleration. Using only Stage 1 causes significant performance degradation—\allowbreak dots.ocr drops from 88.41 to 70.47 on OmniDocBench v1.5—due to the lack of global context and inherited layout segmentation errors from cropped inputs. Thus, 
Stage 2 is essential for recovering the Stage-1-degraded accuracy to the baseline level (88.81 on OmniDocBench v1.5, 92.56 on Ocean-OCR-Bench).
Despite tolerance-based draft verification, the final accuracy remains comparable to the baseline across benchmarks.
These results show that HSD achieves substantial speedups while maintaining parsing quality.

\textbf{Impact of Framework Designs.}
\cref{tab:ablation} ablates the contribution of our framework design on OmniDocBench v1.5 with dots.ocr as the target parser.
Compared with the autoregressive baseline, using a fast pipeline drafter already improves the end-to-end speedup to 2.09$\times$, demonstrating that an efficient drafting stage provides a solid foundation for acceleration.
By introducing the hierarchical two-stage design, our method further increases the end-to-end speedup from 2.09$\times$ to 2.42$\times$ and improves AAL from 2.49 to 3.98, benefiting from region-level parallel verification in Stage 1.
This progression from the baseline to the complete framework demonstrates that the hierarchical design and speculative verification work synergistically to achieve optimal acceleration.

\textbf{Impacts of Drafters.}
In our experiments, we primarily use PP-StructureV3 as the drafter, but our method is flexible and can accommodate other pipelines as drafters. To evaluate the impact of different drafters, we test with MinerU2-pipeline~\cite{MinerUgithub} and Docling~\cite{docling}.\
Additionally, to mimic different real-world drafter behaviors, we add various types of noise to drafts generated by PP-StructureV3, including image degradation, layout detection noise, and recognition errors.
As shown in~\cref{fig:drafter_impact}, despite reduced draft quality, HSD maintains a speedup of over 2.3$\times$, demonstrating its plug-and-play nature and compatibility with various pipelines. This highlights the practical value and robustness of our approach.

\begin{figure}[!t]
  \centering
  \hfill%
  \begin{subfigure}[t]{0.3\linewidth}
    \centering
    \includegraphics[width=\linewidth]{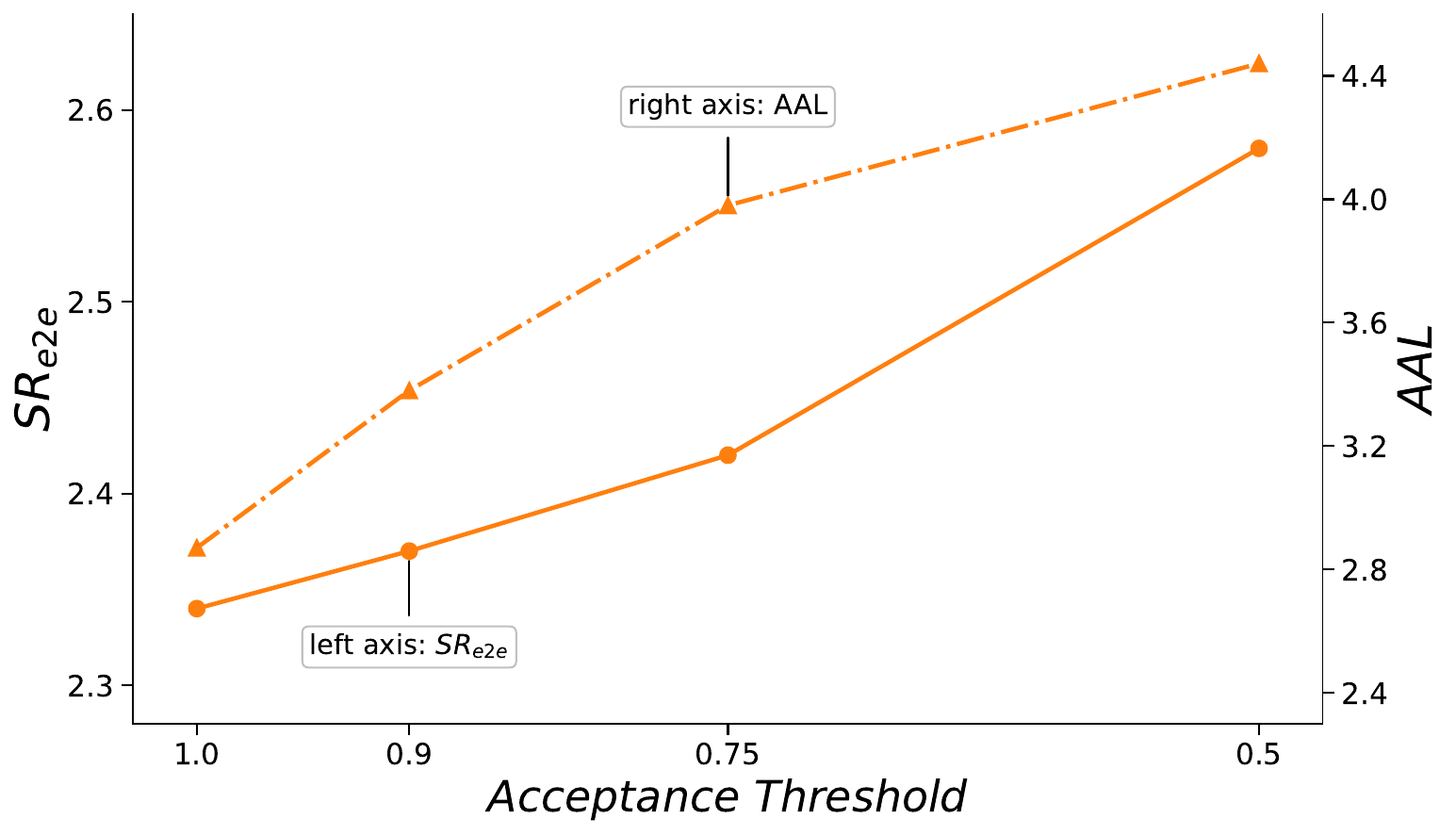}
    \caption{Impact of the \textit{Acceptance Threshold} on $SR_{e2e}$ and \textit{AAL}.}
    \label{fig:acceptance_thresh}
  \end{subfigure}\hfill%
  \begin{subfigure}[t]{0.3\linewidth}
    \centering
    \includegraphics[width=\linewidth]{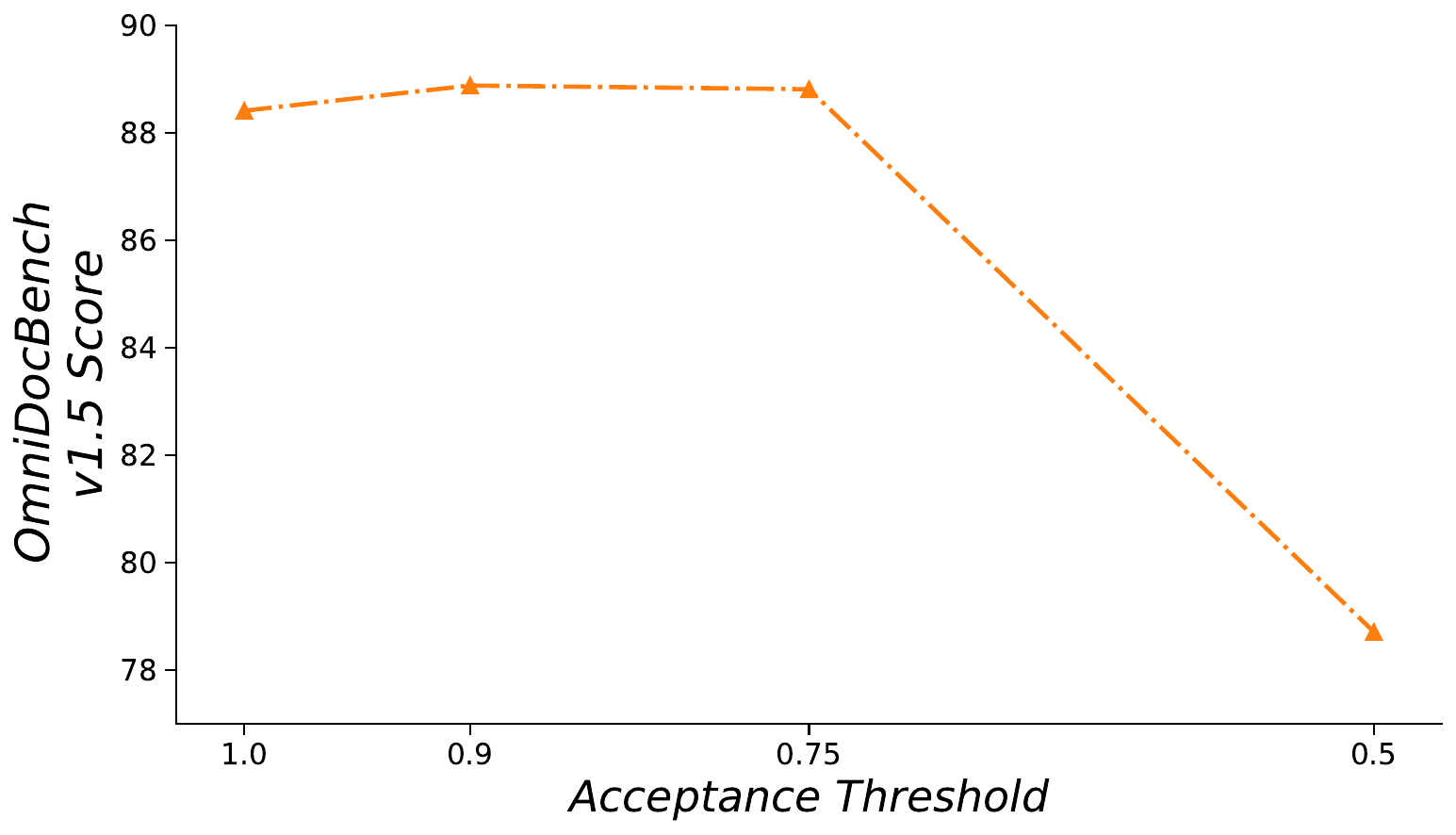}
    \caption{Impact of the \textit{Acceptance Threshold} on parsing accuracy.}
    \label{fig:acceptance_thresh_score}
  \end{subfigure}\hfill%
  \begin{subfigure}[t]{0.3\linewidth}
    \centering
    \includegraphics[width=\linewidth]{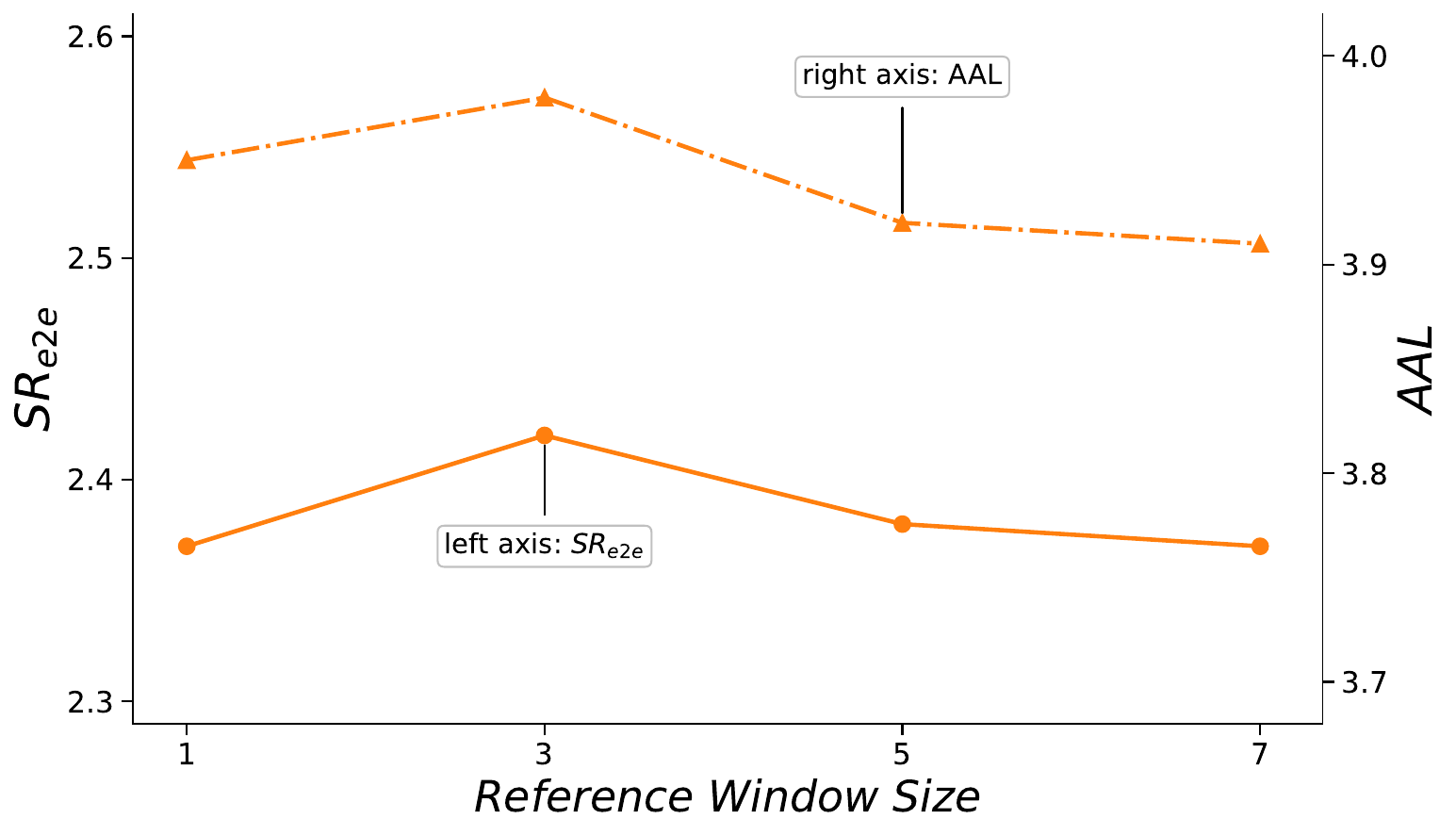}
    \caption{Impact of the \textit{Reference Window Size} on $SR_{e2e}$ and \textit{AAL}.}
    \label{fig:reference_window}
  \end{subfigure}\hfill%

  \caption{Impacts of hyperparameter settings on acceleration and parsing accuracy.}
  \label{fig:hyperparameter}
\end{figure}

\textbf{Impacts of Hyperparameters.}
We further investigate the choices of the acceptance threshold and the reference window size using dots.ocr. As shown in~\cref{fig:hyperparameter}, the acceleration improves as the acceptance threshold decreases. However, when $\tau < 0.75$, the parsing accuracy drops noticeably, while $\tau = 0.75$ achieves the best trade-off between acceleration and accuracy. In addition, the acceleration performance is relatively robust to the reference window size, with $n=3$ yielding the best overall results. Therefore, we set the acceptance threshold $\tau=0.75$ and the reference window size $n=3$.

\section{Conclusion}

In this work, we propose a hierarchical speculative decoding paradigm to address the inference-speed bottleneck of VLM-based end-to-end document parsers. By dividing long, structured document parsing outputs into multiple regions, we enable region-level parallel speculative verification. 
Building on the refined region-level outputs, we perform a full-page speculative verification to preserve the global coherence and correct the remaining errors. 
To further improve the efficiency of HSD, we propose decoupled speculative verification, which resolves draft–target misalignment and enables efficient verification over multiple candidates.
Extensive experiments demonstrate that our approach achieves significant near-lossless speedups: 2.78$\times$ acceleration on OmniDocBench v1.5 with HunyuanOCR and up to 7.04$\times$ acceleration on long-document parsing tasks. These results highlight the practicality and generalizability of our method, offering a plug-and-play solution to accelerate VLM-based document parsers without architectural changes or retraining. We hope that the concept of hierarchical verification will inspire further research in speculative decoding across other fields.

\section*{Acknowledgement}
This research is supported in part by National Natural Science Foundation of China (Grant No. 62476093), the Natural Science Foundation of Guangdong Province (Grant No. 2026A1515012038), Shanghai Artificial Intelligence Laboratory, Guangdong Basic and Applied Basic Research Foundation (Grant No. 2026A1515011967) and Shenzhen Fundamental Research (Key Program) under Grant JCYJ20241202124931042.

\bibliographystyle{splncs04}
\bibliography{main}

\maketitlesupplementary

\setcounter{table}{0}
\renewcommand{\thetable}{A\arabic{table}}
\setcounter{figure}{0}
\renewcommand{\thefigure}{A\arabic{figure}}
\setcounter{section}{0}
\renewcommand{\thesection}{\Alph{section}}

\section{Qualitative Analysis}

\subsection{Qualitative Analysis of HSD Speedup Factors}
\label{sec:qualitative_analysis_speedup}

We qualitatively analyze the factors that dominate the acceleration behavior of our hierarchical speculative decoding (HSD). Specifically, we compare representative pages with \emph{high} end-to-end speedups against those with \emph{limited} speedups using dots.ocr~\cite{dots.ocr}. Across datasets, we observe two recurring bottlenecks for low-speedup cases: (i) \textbf{draft-limited} pages where region/page drafts are inaccurate and thus frequently rejected, and (ii) \textbf{prefill-dominated} pages where the fixed vision/prefill cost accounts for a large portion of the total latency, making decode-side acceleration less visible in end-to-end measurements. Here, \textit{vision/prefill} denotes the front-end stage including the image encoder forward pass and the subsequent multimodal prefill for KV-cache construction.

\begin{figure}[!t]
  \centering
  \includegraphics[width=0.9\textwidth]{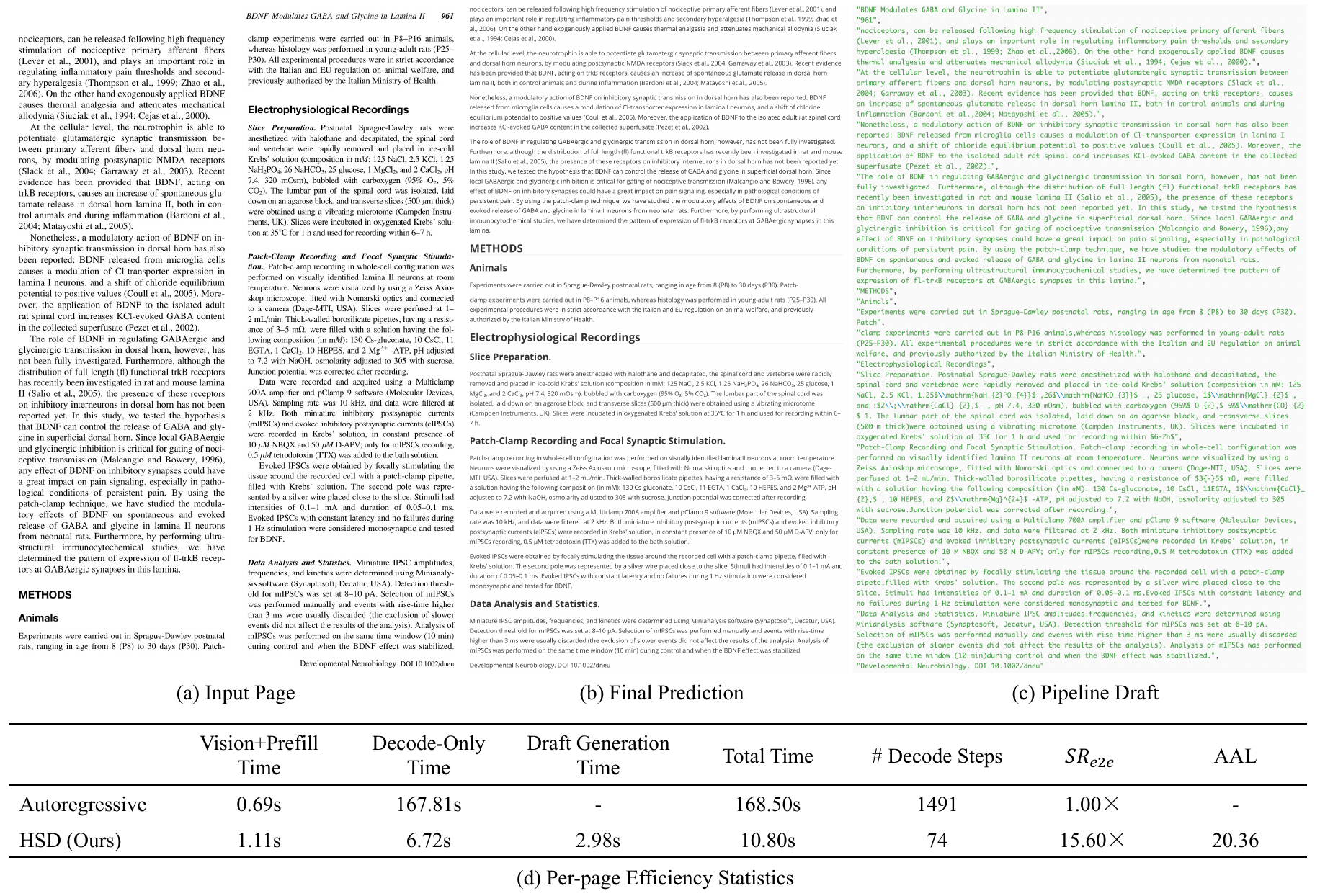}
  \caption{High-speedup example on an English page.}
  \label{fig:qual_high_e}
\end{figure}

\begin{figure}[!t]
  \centering
  \includegraphics[width=0.9\textwidth]{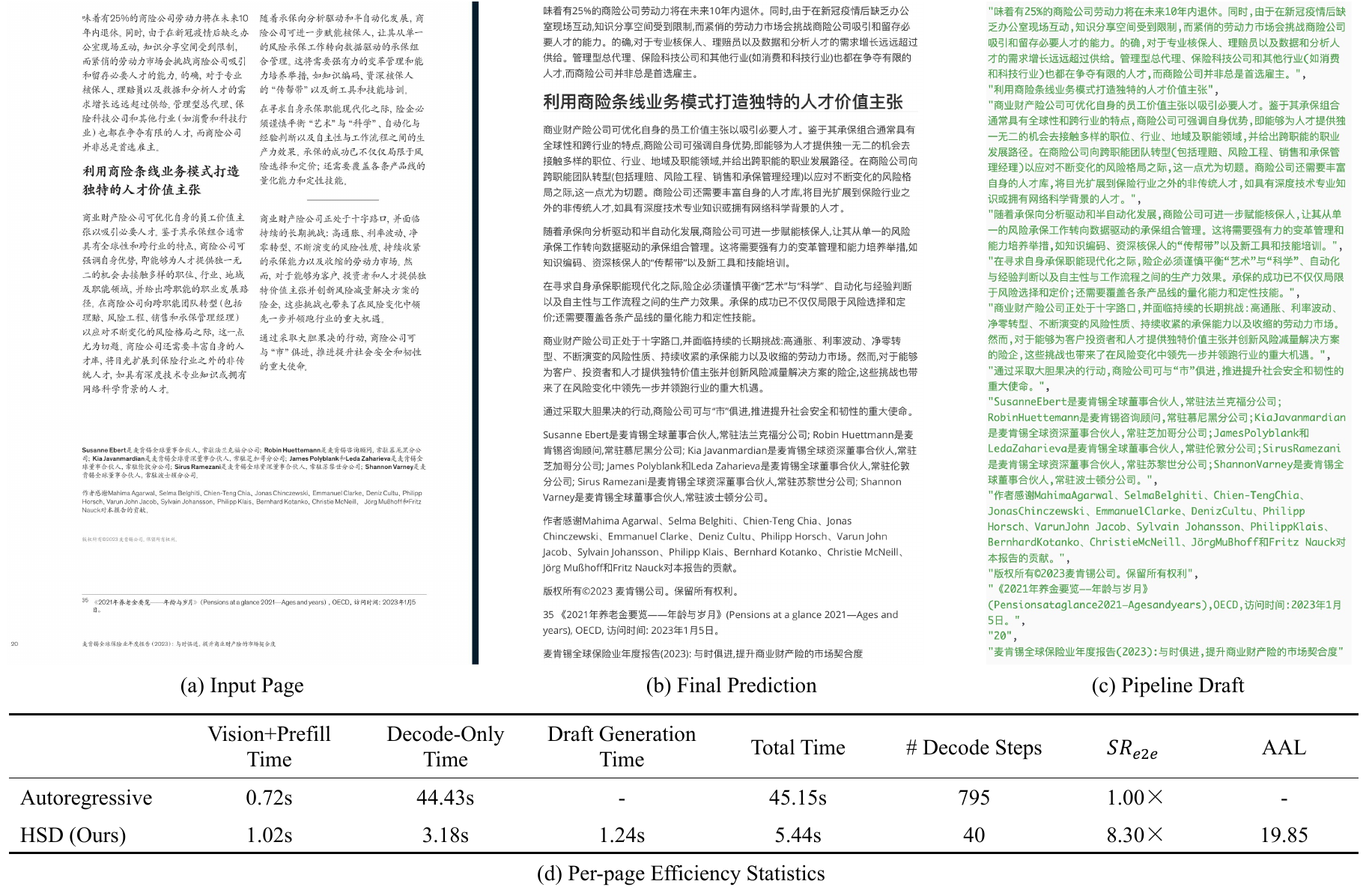}
  \caption{High-speedup example on a Chinese page.}
  \label{fig:qual_high_c}
\end{figure}

\paragraph{High-speedup cases: accurate drafts and decode-dominated latency.}
Figs.\ref{fig:qual_high_e} and \ref{fig:qual_high_c} show typical pages with large speedups (e.g., financial reports, newspapers, or well-structured multi-column layouts). These pages share two properties.
First, the pipeline produces high-quality region and page drafts that closely match the end-to-end parser's output, except for minor formatting variations.
As a result, the end-to-end parser accepts long consecutive draft segments, yielding a high AAL and few rollbacks.
Second, these pages usually contain substantial textual content distributed across many semantic regions, so the decode loop constitutes a major part of the overall latency, while Stage~1 can also leverage richer region-level parallelism.
The fixed vision/prefill cost is amortized over many decoding steps, allowing the decode-side gains to translate directly into strong end-to-end acceleration (high $SR_{\text{e2e}}$).
This indicates that HSD approaches its potential when drafts are reliable and decoding dominates the runtime.

\begin{figure}[!t]
  \centering
  \includegraphics[width=0.75\textwidth]{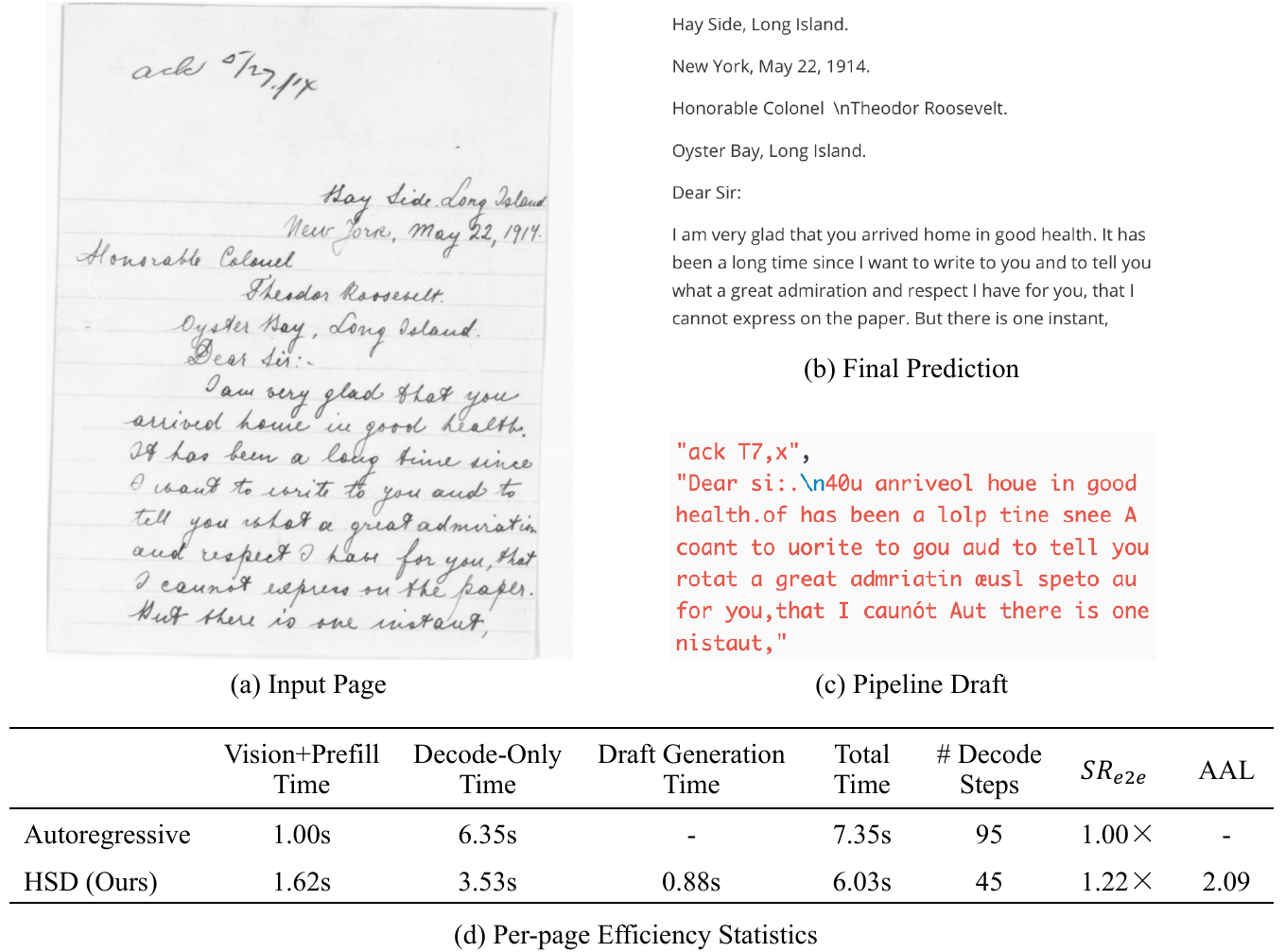}
  \caption{Draft-limited low-speedup example. Draft errors lead to frequent rejections and low AAL.}
  \label{fig:qual_low_draft}
\end{figure}

\paragraph{Low-speedup cases I: draft-limited pages.}
In contrast, \cref{fig:qual_low_draft} shows a representative low-speedup case where acceleration is limited by draft quality. The pipeline draft contains many token-level errors, including missing words and noisy character predictions, which commonly arise when the pipeline struggles with cursive handwriting~\cite{zhang2022msds, dai2023disentangling, dai2024onedm, dai2025beyond}. These errors induce frequent mismatches during verification, resulting in short accepted spans and repeated rollbacks. Consequently, AAL remains low and the end-to-end parser must perform substantial autoregressive corrections, limiting $SR_{\text{e2e}}$.
This case highlights that draft accuracy is a primary factor limiting speculative speedup: when drafts are heavily corrupted, the end-to-end parser cannot reliably accept long segments and thus cannot effectively skip decoding steps.
Improving draft robustness on handwritten content, potentially through large-scale synthetic data generation with recent generative models~\cite{zhang2025ocrgenbench, dai2026guided}, may further alleviate this bottleneck.

\begin{figure}[!t]
  \centering
  \includegraphics[width=0.8\textwidth]{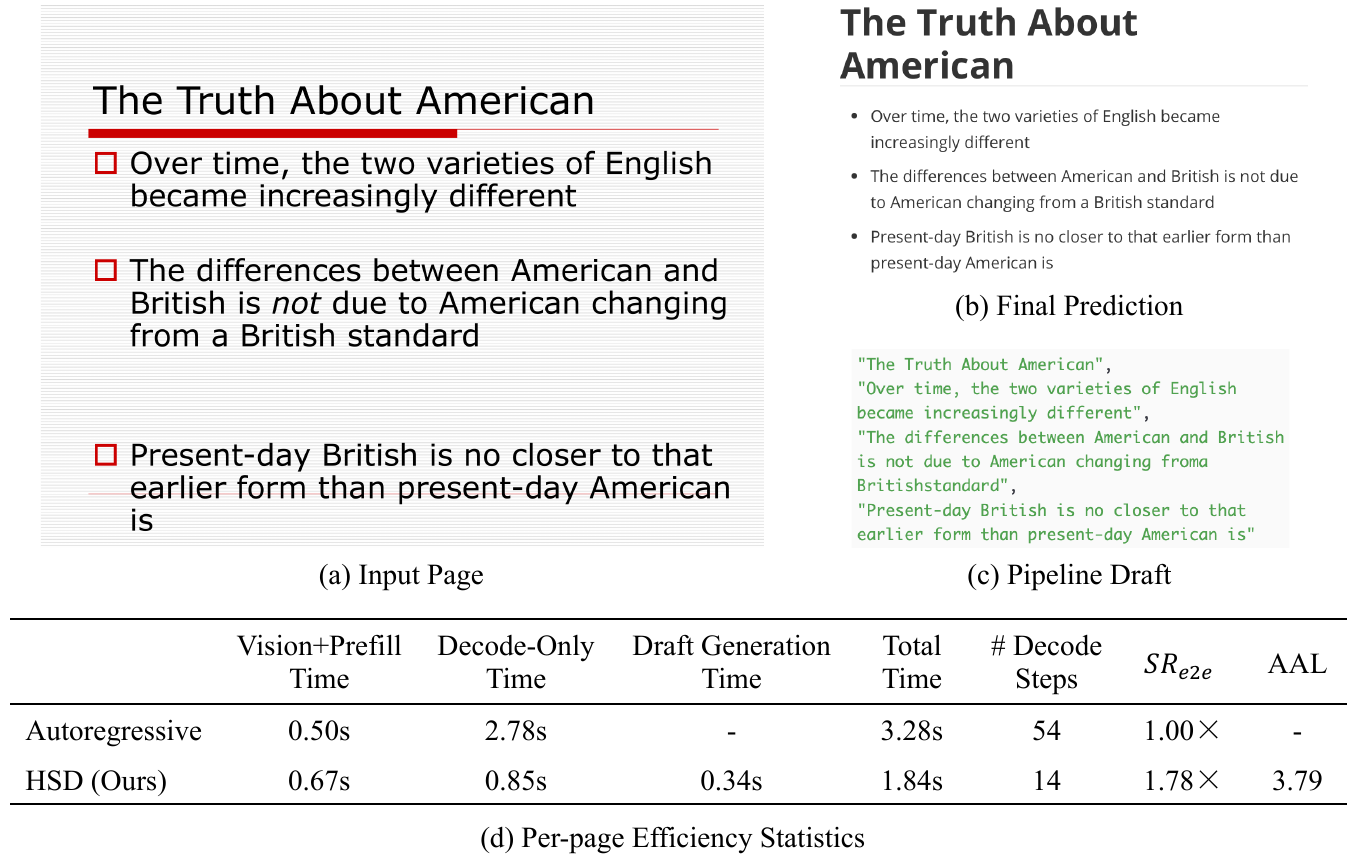}
  \caption{Prefill-dominated low-speedup example. Despite decode-side gains, large fixed vision/prefill cost limits $SR_{\text{e2e}}$.}
  \label{fig:qual_low_prefill}
\end{figure}

\paragraph{Low-speedup cases II: prefill-dominated pages.}

A second failure mode arises for pages where the fixed vision/prefill stage becomes the bottleneck once decoding is accelerated.
As shown in \cref{fig:qual_low_prefill}, such pages typically have short or sparse outputs, leaving limited decoding time to amortize the front-end cost. Even when HSD greatly reduces the decode loop, the overall runtime remains constrained by vision/prefill, so the end-to-end improvement is capped. This explains the smaller gains observed for short documents in our quantitative results: decode-side acceleration alone cannot overcome a front-end cost that occupies a significant fraction of the inference budget.

\begin{figure}[!t]
  \centering
  \includegraphics[width=\textwidth]{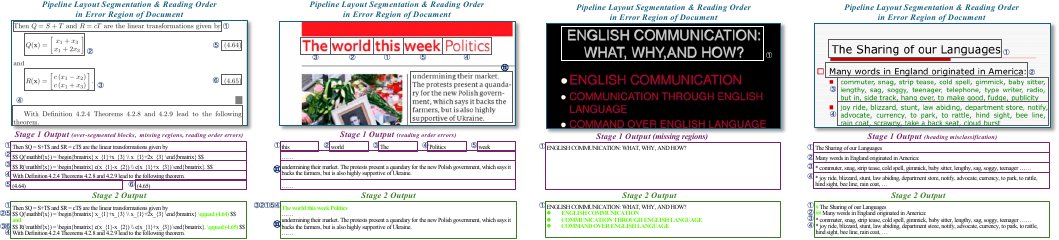}
  \caption{Representative examples of multi-stage error correction in HSD. Stage~1 fixes local errors, whereas Stage~2 corrects remaining structural errors at the page level.}
  \label{fig:vis_stage}
\end{figure}

\paragraph{Takeaways.}
The qualitative evidence supports two key conclusions.
First, draft quality governs AAL and decode-side acceleration: accurate drafts enable long accepted spans and large reductions in decoding steps, whereas draft errors cause frequent rejections and limited gains.
Second, the latency composition governs end-to-end speedup: when the fixed vision/prefill stage accounts for a substantial fraction of total runtime (as in short or sparse pages), decode-side improvements translate only partially into end-to-end speedup ($SR_{\text{e2e}}$).
Taken together, despite variations in the above factors, HSD consistently maintains positive speedup across the analyzed cases. Moreover, it delivers strong overall end-to-end speedup over the whole benchmark reported in our paper, spanning a wide range of document types.

\subsection{Qualitative Analysis of Error Correction Across Stages}
We further analyze representative cases to understand the complementary roles of the two stages in HSD.
As illustrated in~\cref{fig:vis_stage}, Stage~1 is generally effective at resolving local recognition errors in textual content, tables, and formulas, producing drafts close to the final output at the token or element level.
However, it can still leave higher-level structural errors that require global document understanding, such as over-segmented text blocks, missing semantic regions, incorrect reading order, and heading misclassification.
Stage~2 addresses these remaining errors through end-to-end verification and correction.
When the draft is locally accurate but structurally inconsistent, Stage~2 can reject incorrect spans, revise the affected regions, and recover a more coherent page-level representation.

\section{FLOPs and Latency}

Speculative decoding trades increased FLOPs for lower latency because standard autoregressive decoding is typically memory-bound rather than compute-bound.
We characterize this bottleneck with arithmetic intensity (AI)~\cite{williams2009roofline}, defined as FLOPs per byte transferred between GPU high-bandwidth memory (HBM)~\cite{dao2022flashattention} and on-chip compute.
For an NVIDIA A100 80GB SXM GPU, whose peak FP16/BF16 Tensor Core throughput is 312 TFLOPS and HBM bandwidth is 2039 GB/s, the corresponding Roofline ridge point is about 153 FLOPs/byte~\cite{nvidia2021a100datasheet,wang2020timebasedroofline}. Thus, an AI below 153 FLOPs/byte signals a memory bottleneck.
Our evaluation on HunyuanOCR~\cite{hunyuanocr2025tencent} shows that baseline decoding has an AI of just 1.31. Our method boosts the AI to 288, significantly improving compute utilization.
Although FLOPs rise by $41.5\times$, we reduce end-to-end parser forward passes and estimated memory traffic to 0.179$\times$ and 0.189$\times$ of the baseline, respectively, yielding a Roofline-model-predicted 2.81$\times$ speedup.

\begin{table}[!t]
    \caption{Runtime breakdown of HunyuanOCR with and without HSD on valid OmniDocBench v1.5. Pages are grouped by the baseline prefill-time ratio. HSD mainly reduces decoding time, leading to consistent total latency reduction across groups.}
    \label{tab:time_breakdown}
    \centering
    \begin{adjustbox}{width=0.72\linewidth}
      \begin{tabular}{@{}l|c|cc|ccc|cc|c@{}}
        \toprule
        \makecell{Prefill/Full\\Ratio}
        & $N$ 
        & \makecell{Baseline\\Prefill (s)} 
        & \makecell{Baseline\\Decode (s)}
        & \makecell{HSD\\Prefill (s)}
        & \makecell{HSD\\Decode (s)}
        & \makecell{HSD\\Draft (s)}
        & \makecell{Baseline\\Full (s)}
        & \makecell{HSD\\Full (s)}
        & $SR_{e2e}$ \\
        \midrule
        0--2\% & 821  & 0.32 & 44.88 & 0.62 & 11.66 & 2.79 & 45.20 & 15.06 & 3.00$\times$ \\
        2--4\% & 234  & 0.30 & 11.08 & 0.45 & 5.04  & 0.88 & 11.38 & 6.38  & 1.78$\times$ \\
        4--8\% & 134  & 0.29 & 5.26  & 0.44 & 2.69  & 0.81 & 5.56  & 3.94  & 1.41$\times$ \\
        $>$8\% & 151  & 0.28 & 1.81  & 0.38 & 1.19  & 0.42 & 2.09  & 1.99  & 1.05$\times$ \\
        \midrule
        ALL    & 1340 & 0.31 & 30.18 & 0.54 & 8.44  & 1.99 & 30.49 & 10.97 & 2.78$\times$ \\
        \bottomrule
      \end{tabular}
    \end{adjustbox}
\end{table}

\section{Error Accumulation Discussion}

A potential concern is that region-level parallel verification in Stage~1 may introduce error accumulation by making local decisions without full-page context. However, in HSD, such local errors do not accumulate irreversibly, since Stage~2 performs page-level global verification over the aggregated Stage~1 outputs.
By restoring full-page context, Stage~2 can correct residual local errors and cross-region inconsistencies, ensuring that the final output remains globally coherent and faithful to the behavior of the end-to-end parser.
As a result, HSD avoids the cascading errors that may arise in pipeline-based or hybrid parsing methods.

\section{Runtime Breakdown}
\label{sec:runtime_breakdown}

To better understand the source of HSD's latency reduction, we re-profile HunyuanOCR on valid OmniDocBench v1.5 pages and decompose the runtime into prefill and decoding time.
Since different pages exhibit different prefill-decoding cost distributions, we group pages by the baseline prefill-time ratio, defined as
$T_{\mathrm{prefill}}^{AR} / T_{\mathrm{full}}^{AR}$.

As shown in~\cref{tab:time_breakdown}, decoding accounts for a substantial portion of the runtime for most document parsing cases.
HSD mainly reduces the decoding time, with much smaller changes in prefill time, and therefore achieves larger speedups on pages where decoding originally dominates the runtime.
The runtime breakdown further explains why HSD is particularly suitable for document parsing tasks, where outputs are often long and decoding can become the primary latency bottleneck.

\end{document}